\documentclass[11pt, oneside]{Thesis} 

\graphicspath{{Pictures/}} 

\usepackage[round, semicolon, compress]{natbib} 
\hypersetup{hidelinks} 
\title{\ttitle} 

\usepackage{graphicx}
\usepackage{amsmath}
\usepackage{amssymb}
\usepackage{hyperref}
\usepackage{todonotes}
\usepackage{float}
\usepackage[ruled]{algorithm2e}

\begin{document}

\frontmatter 

\setstretch{1.3} 

\fancyhead{} 
\rhead{\thepage} 
\lhead{Chapter \thechapter} 

\pagestyle{fancy} 

\newcommand{\HRule}{\rule{\linewidth}{0.5mm}} 

\hypersetup{pdftitle={\ttitle}}
\hypersetup{pdfsubject=\subjectname}
\hypersetup{pdfauthor=\authornames}
\hypersetup{pdfkeywords=\keywordnames}


\begin{titlepage}
\begin{center}
\includegraphics[width=0.2\textwidth]{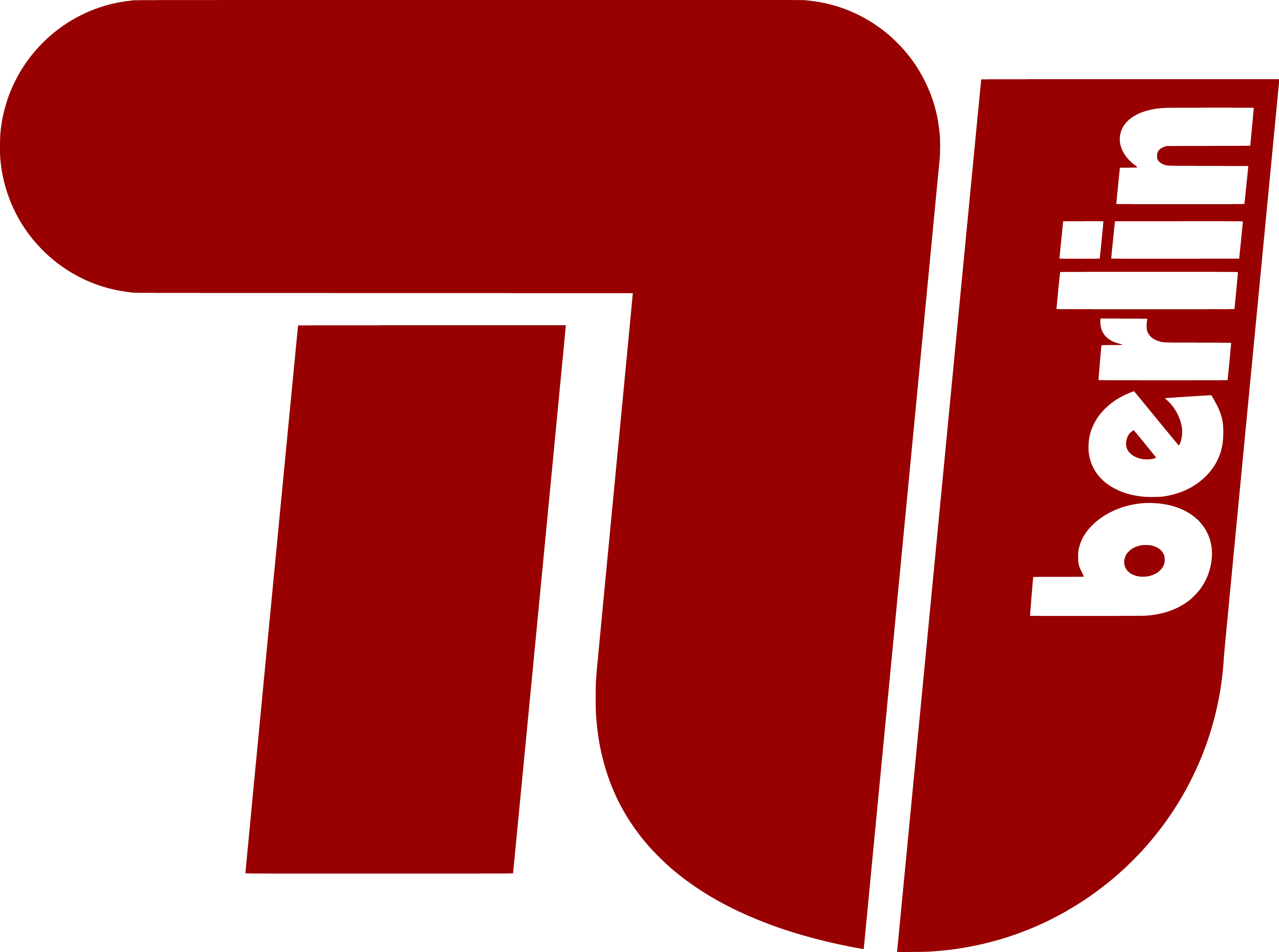}

\textsc{\LARGE \univname}\\[0.5cm] 
\textsc{\Large \facname}\\[1.0cm] 
\textsc{\Large Master Thesis}\\[0.5cm] 

\HRule \\[0.4cm] 
{\huge \bfseries \ttitle}\\[0.2cm] 
\HRule \\[1.2cm] 
 
\begin{minipage}{0.4\textwidth}
\begin{flushleft} \large
\emph{Author:}\\
\authornames 
\end{flushleft}
\end{minipage}
\begin{minipage}{0.4\textwidth}
\begin{flushright} \large
\emph{Supervisors:} \\
\supname 
\end{flushright}
\end{minipage}\\[3cm]
 
\large \textit{A thesis submitted in fulfillment of the requirements\\ for the degree of \degreename}\\[0.3cm] 
\textit{in}\\[0.4cm]
\groupname \\[1cm] 
 {\large \today} \\[1.0cm]

\vfill
\end{center}
\hspace*{-2.0cm}
\includegraphics[scale=0.12]{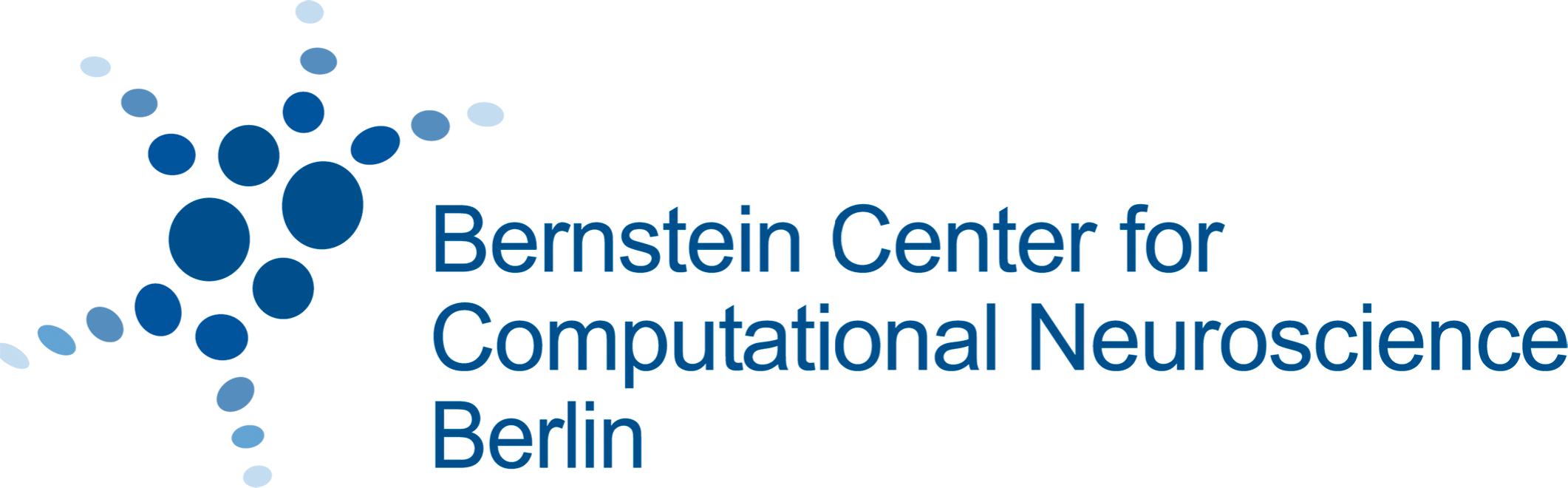}
\hspace*{5.0cm}
\includegraphics[scale=0.17]{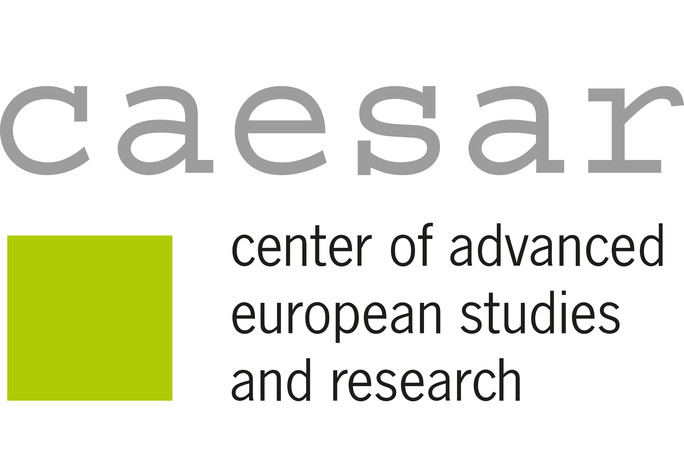}
\end{titlepage}


\pagestyle{empty} 

\null\vfill 

\textit{``It often happens that when two sets of data obtained by observation give slightly different estimates of the true value we wish to know whether the difference is significant. The usual procedure is to say that it is significant if it exceeds a certain rather arbitrary multiple of the standard error; but this is not very satisfactory, and it seems worth while to see whether any precise criterion can be obtained by a thorough application of the theory of probability.''}

\begin{flushright}
Harrold Jeffreys, 1935
\end{flushright}

\vfill\vfill\vfill\vfill\vfill\vfill\null 

\clearpage 


\addtotoc{Abstract} 

\abstract{\addtocontents{toc}{\vspace{1em}} 

A common problem in natural sciences is the comparison of competing models in the light of observed data. Bayesian model comparison provides a statistically sound framework for this comparison based on the evidence each model provides for the data. However, this framework relies on the calculation of likelihood functions which are intractable for most models used in practice. Previous approaches in the field of Approximate Bayesian Computation (ABC) circumvent the evaluation of the likelihood and estimate the model evidence based on rejection sampling, but they are typically computationally intense. Here, I propose a new efficient method to perform Bayesian model comparison in ABC. Based on recent advances in posterior density estimation, the method approximates the posterior over models in parametric form. In particular, I train a mixture-density network to map features of the observed data to the posterior probability of the models. The performance is assessed with two examples. On a tractable model comparison problem, the underlying exact posterior probabilities are predicted accurately. In a use-case scenario from computational neuroscience -- the comparison between two ion channel models -- the underlying ground-truth model is reliably assigned a high posterior probability. Overall, the method provides a new efficient way to perform Bayesian model comparison on complex biophysical models independent of the model architecture. \\[4.5cm]
}

\clearpage 

\addtotoc{Deutsche Zusammenfassung} 

\zusammenfassung{\addtocontents{toc}{\vspace{1em}} 
Ein häufiges Problem in den Naturwissenschaften ist der Vergleich konkurrierender Modelle im Hinblick auf beobachtete Daten. Der Bayes'sche Modellvergleich liefert einen statistisch fundierten Rahmen für diesen Vergleich, basierend auf der Evidenz eines jeden Modells für die Daten. Dieses Verfahren beruht jedoch auf der Berechnung von Likelihood-Funktionen, die für die meisten in der Praxis verwendeten Modelle nicht definiert sind. Bisherige Ansätze auf dem Gebiet der Approximativen Bayes’schen Berechnung (ABC) umgehen die Berechnung der Likelihood-Funktion und schätzen die Evidenz eines Modells basierend auf der sogenannten Verwerfungsmethode, sie sind jedoch typischerweise rechenintensiv. In dieser Arbeit schlage ich eine neue effiziente Methode vor, um den Bayes'schen Modellvergleich im Rahmen von ABC durchzuführen. Dabei stütze ich mich auf jüngste Fortschritte in der Dichteschätzung und approximiere die A-Posteriori-Wahrscheinlichkeitsverteilung über Modelle in parametrischer Form. Genauer, ich trainiere ein künstliches neuronales Netzwerk, um Eigenschaften der beobachteten Daten auf die A-Posteriori-Wahrscheinlichkeit eines jeden Modells abzubilden. Die Performanz der Method wird anhand von zwei Beispielen demonstriert. In einem berechenbaren Modellvergleich Problem wird die zugrunde liegende exakte A-Posteriori-Wahrscheinlichkeit akkurat vorhergesagt. In einem Anwendungsszenario aus dem Bereich der theoretischen Neurowissenschaften -- der Vergleich zweier Ionenkanalmodelle -- wird dem zugrunde liegenden Modell eine hohe A-Posteriori Wahrscheinlichkeit zugeordnet. Insgesamt bietet die neue Methode eine effiziente Möglichkeit, den Bayesschen Modellvergleich an komplexen biophysikalischen Modellen durchzuführen, unabhängig von der Modellarchitektur.  \\[2cm]
}

\clearpage 


%
%
%
%


\pagestyle{fancy} 

\lhead{\emph{Contents}} 
\tableofcontents 


\mainmatter 

\pagestyle{fancy} 
\lhead{Chapter \thechapter} 


\chapter{Introduction}
\label{ch:introduction}
The goal of natural and social sciences is to acquire and organize knowledge by providing explanations and predictions about the world. Explanations and predictions are often based on hypotheses and models developed from observations and from prior knowledge. Based on this general approach, scientists often develop multiple hypotheses and models based on the set of observations. Therefore, one essential step in the scientific process is the comparison of models when given observations. 

The field of statistics has developed many techniques for model comparison. One common technique, especially in the psychological and social sciences, is null hypothesis significance testing (NHST) based on \textit{p}-values \citep{neyman1933}. Although commonly used, this technique has been subject to criticism for numerous years. Among the main concerns are the following. First, the meaning of a significant result is often misunderstood or misinterpreted by researchers \citep{nickerson2000null, wasserstein2016asa}. Second, the test only quantifies the evidence under the null hypothesis and rejects it if it is low, irrespective of the evidence under the hypothesis of interest \citep{wagenmakers2017need}. As an alternative technique, critics commonly suggest hypothesis testing based on Bayesian model comparison, as it is easier to interpret and theoretically sound \citep{dienes2011bayesian}.

In Bayesian model comparison, competing statistical models are not compared against a null hypothesis but among each other, on the basis of the evidence they provide for the observed data. The result of the comparison is a posterior probability for each model, i.e. the conditional probability that is assigned to a model given the observed data. The rationale of this approach lays in Bayes' theorem. It states that the posterior probability of a model is a consequence of two terms: a prior probability that expresses prior beliefs, and a likelihood function expressing the probability of the data under the model, i.e. the model evidence. In situations where the prior beliefs about the models are indifferent, models can be compared based on the Bayes factor \citep{jeffreys1935} that is given by the ratio of the model evidences. 

The model evidence is therefore the central term in Bayesian model comparison. It can be interpreted as the likelihood of the data under the model with all model parameters marginalized out. It therefore implicitly penalizes overly complex or too simple models and is less sensitive to overfitting. However, this step requires that the likelihood function for the model is available. It thereby brings about one of the major limitations of the practical uses for Bayesian model comparison \citep{vandekerckhove2015model}: the likelihood function, and therefore the model evidence, is difficult or impossible to calculate for many scientific models used in practice. This applies especially to areas like physics, biology or neuroscience, where models are often defined as differential equations. Accordingly, the field of approximate Bayesian computation (ABC) \citep[reviewed in][]{sunnaaker2013approximate} has emerged with diverse contributions from these areas by aiming to apply Bayesian methods to statistical models with intractable likelihood functions \citep{rubin1984}.

One of the main ideas in ABC is to circumvent the evaluation of the likelihood function by generating data from a model and comparing it to the observed data. This approach can be used for both the estimation of the parameters of a model and for model comparison. In case of model comparison, the model evidence is approximated by testing how well the data generated from one model can mimic the observed data. This was initially proposed by \cite{rubin1984} as the basic rejection sampling approach for model comparison: data is generated for every model and for a large range of parameters, which is then compared to the observed data by using a distance measure. If the distance is below some threshold, the simulation is accepted, otherwise it is rejected. The relative number of accepted simulations of a model then provides an estimate of the model evidence. 

A major drawback of this method is the trade-off between accuracy and the number of simulations: The smaller the acceptance threshold, the better the estimate; alas, a smaller threshold also requires more simulations. Several improvements were adjoined to rejection sampling in order to enhance sampling efficiency and accuracy \citep{toni2009, leuenberger2010, didelot2011}. However, for models with many parameters, the number of needed simulations still becomes impractically high. In addition, the inherent problem of choosing an appropriate distance measure on the observed data remains. 

The method of rejection sampling originated from the problem of estimating the posterior distribution over parameters of a model \citep{marjoram2003, toni2010}. In this subfield of parameter estimation within ABC, several approaches have been presented to overcome its limitations. One crucial idea for avoiding rejection sampling was proposed by \cite{beaumont2002} who estimated the properties of posterior distributions, such as the mean or the probability density curve, with a regression from model parameters onto the corresponding simulated data. This idea was developed further by \cite{blum2010}, who introduced non-linear regression approaches for posterior estimation, and by \cite{papamakarios2016}, who used Bayesian conditional density estimation to estimate the posterior. 

In general, these alternative methods estimate a parametric form of the posterior, instead of approximating it with a set of accepted samples. For example, in \cite{papamakarios2016} the parametric form of the posterior is approximated by training an artificial neural network to learn a mapping from simulated data and parameters to a posterior over parameters. The training data set consists of pairs of parameters from a prior distribution and corresponding data generated from the model. This approach entails several advantages over rejection sampling: all simulated data is used, no distance measure has to be chosen, and a closed form posterior distribution is obtained. Recently, \cite{lueckmann2017} refined this approach further and demonstrated its applicability for common neuroscience models with a large number parameters and complex prior distributions. 

So far there has been no attempt to extend the promising technique of posterior density estimation in ABC to the estimation of the posterior distribution over models. This thesis presents a method for Bayesian model comparison in approximate Bayesian computation based on posterior density estimation, building on the work by \citet{papamakarios2016} and \citet{lueckmann2017}. The goal of this new method is the following: Given a set of models for which the likelihoods are intractable but which can be simulated to generate data, and given some observed data, estimate the posterior probability of each model given the observed data. Based on this posterior probability, Bayesian model comparison can be performed, e.g. using the Bayes factor.

The method is theoretically grounded on estimating the density function of the joint probability distribution of models, model parameters, and data. Therefore, following the procedure in \citet{papamakarios2016}, it is possible to derive a loss function that guarantees convergence of the estimated posterior to the true posterior in the limit of infinite samples. This loss function is used to train the neural network that approximates the mapping from observed data to both the posterior over models and the posterior model parameters. 

To demonstrate the feasibility of this approach, I will test it on a theoretical example of model comparison for which the ground-truth posterior over models is known, and on an unidealized use-case scenario for which the ground-truth posterior is not known. Both example problems are taken from the field of computational neuroscience. 

The theoretical example is the comparison between a Poisson model and a negative-binomial model. For both models, it is possible to calculate the true posterior probability so that the proposed method can be validated directly. Furthermore, the comparison is relevant from a neuroscientific perspective. Both models are used to mirror the statistics of neural spiking data, in particular the variability of spike counts between trials of an experiment. Usually, spike count variability is modeled as a Poisson process, assuming equal mean and variance in the counts, as reflected in cortical responses to visual stimuli \citep{shadlen1998, mcadams1999}. However, it has also been reported that the spike count variability is higher than expected from a Poisson model \citep{dean1981, vogels1989}, implying that the spike counts are overdispersed. \cite{taouali2015} therefore proposed to assume a negative-binomial model instead, offering an additional degree of freedom on the variance parameter which allows the variance to be larger than the mean. 
I will use surrogate data from these models in order to perform Bayesian model comparison. 

The second example is the comparison of two biophysical models of ion channels based on simulated current traces. This mimics a common problem in computational neuroscience: the choice of the appropriate computational model for data that was observed experimentally. To simplify the systematic comparison of published ion channel models \cite{podlaski2017} developed an ion channel genealogy (ICGenealogy) that compares ion channel models based on publication metadata and on the responses of the ion channel models to stereotypical experimental paradigms. For the example scenario, I select two ion channel models, both from \cite{pospischil2008} and available in the ICGenealogy, and perform Bayesian model comparison between them based on the stereotypical experimental paradigms proposed by \cite{podlaski2017}. \\[0.5mm]


In summary, this thesis presents a new method for Bayesian model comparison for models with intractable likelihoods. In contrast to previous methods that estimate the model evidence using rejection sampling, this method obtains the posterior over models using conditional density estimation. The procedure is tested on an idealized comparison between a Poisson model and a negative-binomial model for which the exact posterior probabilities can be calculated, and on a realistic use-case scenario of comparing two ion channel models.

The thesis is structured in six chapters. After this introduction, the second chapter will provide a more detailed introduction to Bayesian model comparison, an overview of related work in approximate Bayesian computation, and a motivation for the proposed method. The third chapter will present the methods for model comparison through posterior density estimation. Chapters four and five contain the methods and results of the theoretical example and the use-case example, respectively. Finally, chapter six will discuss the results. 
\chapter{Theoretical Background}
\label{ch:background}
The goal of this chapter is to provide a more detailed introduction to the methods and concepts used in this thesis. I begin with an introduction to Bayesian model comparison and an outline of how it is approached in the field of approximate Bayesian computation (ABC).  A lot of work in ABC focuses on estimating the posterior over the parameters of a model. I will therefore first introduce rejection sampling and posterior density estimation approaches from this sub-field of ABC, before addressing model comparison approaches. In the last section of this chapter I will comment on problems that arise when using summary statistics of the data for model comparison in ABC. 

\section{Bayesian model comparison}
\label{sec:background_bmc}
The goal of the Bayesian model comparison is to obtain a posterior probability for each model using Bayes' rule. Given a set of model classes $\mathcal{M} = \{\mathcal{M}_i \ldots \mathcal{M}_n \}$ and data $\mathcal{D}$ the posterior over models is given by 
\begin{align}
p(\mathcal{M} | \mathcal{D}) = \frac{p(\mathcal{D}| \mathcal{M}) p(\mathcal{M})}{p(\mathcal{D})}.
\end{align}
$p(\mathcal{M})$ is the prior distribution over model classes. Assuming that the data $\mathcal{D}$ is generated from one of the models, the prior expresses our uncertainty about which one it is. It is chosen by the experimenter and can be used to express a preference for a certain model. 
 
The term $p(\mathcal{D}| \mathcal{M})$ is the model evidence, often called the marginal likelihood. This term relates data and models. For a model $\mathcal{M}_i$ it is given by 
\begin{align}
p(\mathcal{D}| \mathcal{M}_i) = \int p(\mathcal{D}| \theta, \mathcal{M}_i) p(\theta | \mathcal{M}_i) d\theta, 
\end{align}
where $\theta$ are the parameters of the model $\mathcal{M}_i$. It is the probability of the observed data $\mathcal{D}$ given a particular set of model parameters $\theta$, weighted with the corresponding prior probability of $\theta$, integrated over all parameters $\theta$. Thus, it is a marginalization over the parameters of the model. This is an important point as it implies that overfitting is avoided so that in Bayesian model comparison models can be compared directly on the training data \citep{bishop2006}. 

Given this definition of the posterior over models, a model can be selected based on the maximal posterior probability. A related important quantity for comparing models is the \textit{Bayes factor} \citep{kass1995}. The Bayes factor is defined as the ratio of the model evidences
\begin{align}
B_{12} &= \frac{p(D | \mathcal{M}_1)}{p(D | \mathcal{M}_2)} = \frac{\int p(\mathcal{D}| \theta, \mathcal{M}_1) p(\theta | \mathcal{M}_1) d \theta}{\int p(\mathcal{D}| \theta, \mathcal{M}_2) p(\theta | \mathcal{M}_2) d \theta}
\end{align}
expressing the evidence provided by the data in favor of one model over another. A Bayes factor much larger than 1 indicates strong evidence for model $\mathcal{M}_1$ whereas a Bayes factor close to 0 indicates evidence is in favor of $\mathcal{M}_2$. 

The Bayes factor is often proposed as an alternative to hypothesis testing \citep{wagenmakers2007} because it has several advantages. First, the models in question do not need to be related in any way, e.g. they need not to be nested. Second, the Bayes factor provides evidence in favor of and against a model, unlike in hypothesis testing where a null hypothesis that was not rejected allows no further conclusion. And third, the Bayes factor between two models gives an intuitive measure for the weight of evidence for the model. 

Several approaches have been made to give an interpretation to the weight of evidence associated with a certain value of the Bayes factor \citep{jeffreys1935, kass1995}. This results in suggestions for when to accept one or another hypothesis, similar to levels of significance in null-hypothesis significance testing. However, the Bayes factor is directly related to the ratio of the posterior probabilities: 
\begin{align}
\label{eq:bayes_factor_posterior_odds}
B_{12} &= \frac{p(\mathcal{D} | \mathcal{M}_1)}{p(\mathcal{D} | \mathcal{M}_2)} = \frac{p(\mathcal{M}_1 | \mathcal{D})p(\mathcal{M}_2) }{p(\mathcal{M}_2 | \mathcal{D})p(\mathcal{M}_1)}, 
\end{align}
and the posterior probabilities give an intuitive interpretation of the weight of evidence of one model over another. In the following I will therefore compare models based on their posterior probabilities. Note that if the prior distribution over models is uniform, the Bayes factor is equal to the ratio of posterior probabilities. 

\section{Approximate Bayesian computation}
The framework of Bayesian model comparison comes with one limitation, that is, the difficulty of calculating the model evidence in practice. For most models used in the scientific context the likelihood is not defined or costly to evaluate, and therefore, the integral of the likelihood over all model parameters is usually not tractable. The field of approximate Bayesian computation (ABC) develops methods for this scenario. 

In a situation where the likelihood $p(\mathcal{D} | \theta)$ of a model is not tractable, it is often still possible to sample from the joint distribution of parameters and data, $p(\mathcal{D}, \theta)$. This is the case when the model is defined as a simulator that generates data $x_o \in \mathcal{D}$ given a set of parameters $\theta$, i.e. it samples from $p(\mathcal{D} | \theta)$. Sampling a parameter from a prior $p(\theta)$ and simulating the model to obtain data $x_o$ then corresponds to a sample from the joint distribution $p(\mathcal{D}, \theta)$. Therefore, it is possible to approximate the posterior by sampling parameters and simulating data without explicitly evaluating the likelihood. 

Many methods developed in the field of ABC are designed to approximate the posterior over parameters, $p(\theta | \mathcal{D})$, given some observed data $\mathcal{D}=x_o$, a generative model with parameters $\theta$ and a prior $p(\theta)$. This is a task that is not directly related to model comparison because it deals with finding the parameters of one particular model. However, the methods for approximating the posterior over models are based on these methods. Therefore, I will explain them first, before turning to the problem of approximating the posterior over models. The approaches can be roughly separated into rejection sampling approaches that approximate the posterior with a set of samples and density approaches that learn a parametric form of the posterior. 

\subsection{Rejection sampling approaches}
\label{sec:rejection_sampling}
The central idea of most ABC methods is to replace the calculation of the likelihood $p(\mathcal{D} | \theta)$ by a comparison of the simulated and the observed data. Given observed data $\mathcal{D}=x_o$, a simulator model $f(x | \theta)$ and a prior over model parameters $p(\theta)$, the posterior is approximated by a set of parameter samples that yield simulated data similar to the observed data. The most basic approach in this direction is the \textit{ABC rejection sampler} \citep{pritchard1999}:

\begin{algorithm}[H]
\SetAlgoLined
Given: a prior $p(\theta)$, a simulator $f(\theta)$, observed data $x_o \in \mathcal{D}$\;
Set number of total simulations $N$\;
 \For{$i=1$ \KwTo $N$}{
 Sample parameters $\theta_i$ from prior $p(\theta)$\;
 Simulate a data set $x_i$ from the simulator $f(x|\theta_i)$\;
  \eIf{$d(x_o, x_i) \leq \epsilon$}{
   accept $\theta_i$\;
   }{
   reject $\theta_i$\;
   }
 }
 \caption{ABC rejection sampler}
 \label{algo:rejection_sampler}
\end{algorithm}

Here, $d(x, x')$ is some distance measure between data $x$ and $x'$ and $\epsilon$ is a tolerance level. The result of the sampling algorithm is a set of samples from the distribution $p(\theta | d(x_o, x_i) \leq \epsilon)$ that for sufficiently small tolerance $\epsilon$ will approximate the posterior distribution $p(\theta | x_o)$ well. However, the acceptance criterion $\epsilon$ constitutes one of the main problems of this algorithm. With a prior on the parameters that is broad in comparison to the true posterior and a low tolerance $\epsilon$ the acceptance rate will be very low so that a very large number of samples is needed. 

To overcome this problem, several improved versions of rejection sampling have been proposed, usually based on more systematic sampling methods like Markov Chain Monte Carlo (MCMC) \citep{marjoram2003}. As an example, I give a short outline of the sequential Monte Carlo (SMC) algorithm \citep{sisson2007} with importance sampling proposed by \cite{toni2009} that combines several of the improvements. 

The idea is to repeat the rejection sampling algorithm presented above multiple times. To gradually approach the target distribution of the posterior and to sample more efficiently, the tolerance $\epsilon$ is adapted to a smaller value in every round dependent on the number of accepted samples. Additionally, new parameter samples are drawn from a proposal distribution instead of the prior. The proposal distribution is usually chosen as the set of accepted samples from the previous round and additionally, the samples selected from the proposal distribution are perturbed with a kernel to change them slightly. Furthermore, for each accepted parameter sample an importance weight is calculated to capture the importance of that sample for the approximation of the posterior \citep{del2006}, e.g. by taking into account the probability of the sample under the proposal distribution. The final estimate of the posterior is then given by the set of accepted samples and their importance weights. 

All sampling based approaches require the choice of a distance measure $d$ and a tolerance $\epsilon$. This implies a trade-off between accuracy and simulation time because the posterior is approximated well only for small $\epsilon$ and the acceptance rate is proportional to $\epsilon$. As $\epsilon$ is reduced to obtain a better estimate, the number of required simulations becomes intractable for models with many parameters. Another disadvantage is given by the fact that the approximated posterior is not given in closed form but by a set of samples, which complicates the interpretation of the shape of the posterior and the comparison to other posteriors. These disadvantages can be overcome by parametric approaches based on density estimation. 

\subsection{Density estimation approaches}
The goal of posterior density estimation in ABC is to learn a parametric approximation to the exact posterior. This goes back to \cite{beaumont2002} who proposed to learn properties of the posterior, such as its mean and variance, directly using a regression from simulated parameters to generated data. More recently, this idea was developed further by \cite{papamakarios2016} who proposed mixture-density networks (MDN) \citep{bishop1994} to estimate the entire conditional density of the parameters given the observed data $p(\theta | \mathcal{D}=x_o)$. 

The rational behind MDN is to learn a mixture model $q_{\phi}(\theta | x)$ that for a given input data $x$ approximates any conditional density $p(\theta | x)$. Once the mixture model is trained, it can in principle predict the posterior for many different data points of interest. As in \cite{papamakarios2016} the model $q_{\phi}$ could be a mixture of Gaussians so that the parameters of the MDN $\phi$ are the means, variances and the mixture coefficients: 
\begin{align}
q_{\phi}(\theta | x) &= \sum_{k=1}^K \alpha_k(x) \mathcal{N}(\theta | \mathbf{\mu}_k(x), \sigma^2_k(x)). 
\end{align}
The dependence between parameters $\phi=\{\alpha, \mu, \sigma \}$ and the data $x$ is expressed with a conventional artificial neural network that takes data $x$ as input and produces the means, variances and mixture coefficient for every mixture component as output. To approximate the posterior over the parameters given observed data $x_o$ the MDN is trained with samples $(\theta, x)$ generated from the prior $p(\theta)$ and the simulator model. A prediction is obtained by evaluating the optimized MDN at the observed data $\mathcal{D} = x_o$. 

\cite{papamakarios2016} learn the posterior with several improvements to the basic MDN setup. They replace the training of the classic MDN by a Bayesian version using stochastic variational inference \citep{kingma2013}. Then, similar to the SMC approach, they replace the original prior over parameters $p(\theta)$ by a proposal distribution $\tilde{p}(\theta)$ and adapt the proposal distributions over multiple rounds of training. In every round parameters are drawn from the proposal prior and the model is simulated to obtain samples $(\theta, x)$. The samples are used to train the MDN for a first approximation of the posterior. In turn, this approximation is used as the proposal distribution for next round. 

This method was demonstrated to work well on theoretical and practical examples and to require less samples than Monte Carlo ABC methods. However, it entails some limitations too. First, the choice of the proposal distribution is restricted to uniform and Gaussian distributions. Second, the replacement of the original prior by a proposal distribution requires a post-hoc correction to actually approximate the original posterior and not the posterior based on the proposal distribution, which caused numerical instabilities in some cases. \cite{lueckmann2017} addressed these issues. They provide an importance-weighted loss function that directly corrects for the use of complex proposal priors. Additionally, they present an approach to effectively deal with simulations that result in missing values and demonstrate that the method works well on typical biophysical models. 

Overall, posterior density estimation approaches are a promising alternative to rejection sampling as they seem to achieve similar or better results \citep{lueckmann2017}, yield parametric posterior estimates and seem to scale better to models with many parameters.

After this short overview of approaches to posterior estimation in ABC, I will now turn to the actual problem of approximating the posterior over models. 

\subsection{Model comparison approaches}
Bayesian model comparison requires the calculation of the posterior over models or the model evidences. This calculation becomes a subject of ABC as it relies on the evaluation of the likelihood of the model which is often not tractable. Most approaches to model comparison in ABC are based on rejection sampling. The basic form of the rejection sampling algorithm as illustrated above (see section \ref{sec:rejection_sampling}) was extended to model comparison by \cite{rubin1984}. Given a discrete random variable $m$ over the set of models $\mathcal{M}$, a prior over models $p(m)$, the prior over parameters for every model $p(\theta | m)$ and the simulators, the model evidence is approximated by the relative frequency with which a model generates data similar to the observed data. The procedure is shown in algorithm \ref{algo:rejection_sampler_model_comparison}: \\ 
\begin{algorithm}[H]
\SetAlgoLined
Given: priors $p(m)$ and $p(\theta | m)$, a simulator $f(\theta, m)$, observed data $x_o \in \mathcal{D}$\;
Set number of total simulations $N$\;
 \For{$i=1$ \KwTo $N$}{
 Sample a model $m_i$ from the prior $p(m)$\;
 Sample parameters $\theta_i$ from $p(\theta | m_i)$\;
 Simulate a data set $x_i$ from $f(x|\theta_i, m_i)$\;
  \eIf{$d(x_o, x_i) \leq \epsilon$}{
   accept $(m_i, \theta_i)$\;
   }{
   reject $(m_i, \theta_i)$\;
   }
 }
 \caption{ABC rejection sampler for model comparison}
 \label{algo:rejection_sampler_model_comparison}
\end{algorithm}
 
After convergence according to a criterion on the number of simulations or on the tolerance $\epsilon$, the posterior probability of model $m_i$ is estimated as 
\begin{align}
p(m_i | x_o) \approx \frac{\text{\# accepted samples for } m_i}{N}, 
\end{align}
where $N$ is the number of total simulations. That is, the more often a sample belonging to a model $m_i$ is accepted, the higher the evidence for this model. With this estimate of the posterior probability the Bayes factor between individual models can be calculated. 

This rejection based algorithm suffers from the same limitations as mentioned above: the sampling procedure becomes inefficient if the prior is broad and the tolerance is small. As an improvement, \cite{toni2009} extended the SMC approach presented above to the scenario of model comparison. In this extension the models are represented by an additional model parameter for which the posterior is estimated as the relative frequency of acceptance of the parameters of a model, similar to the basic rejection sampling approach. However, the procedure of sampling new parameters follows the more efficient SMC approach over several rounds with adaptive tolerance, proposal distributions and importance weights. Similarly, \cite{didelot2011} use a SMC approach to estimate the model evidence, however, they estimate the evidence independently for every model. \cite{leuenberger2010} use rejection sampling paired with regression adjustment \citep{beaumont2002} to allow for larger tolerance values. 

A technique different from rejection sampling was proposed by \citet{pudlo2015} who phrase the model comparison as a classification problem that is approached with the random forest algorithm \citep{breiman2001}. In a first step a random forest classifier is applied to the data to predict the model that fits best and in a second step the posterior probability of that model is approximated with a regression relying again on the random forest algorithm. This seems to be computational more efficient compared to sampling approaches. However, it remains unclear how the classification approach via random forest relates to the exact posterior probability. 

In summary, most approaches to the estimation of the posterior over models rely on rejection sampling and its variants. The literature presented above mostly originates in field of genetics and computational biology and the developed methods have been demonstrated to work on corresponding examples. Accordingly, several software packages for performing Bayesian model comparison on problems in this field using rejection sampling are publicly available \citep{mertens2018, cornuet2014, liepe2014}. 

However, being based on rejection sampling these methods come with the drawback of depending on the choice of a distance measure and a tolerance level. Additionally, they usually do not scale to higher-dimensional parameter spaces. As mentioned above, parametric approaches have been proposed to overcome these limitations for the estimation of the posterior over parameters of a model. Chapter \ref{ch:general_methods} will present the extension of these approaches to the estimation of the posterior over models. 

\subsection{Summary statistics}
\label{sec:summary_stats}
One important point that has not been addressed concerns the summary statistics of the data. The methods as illustrated above seem to operate directly on the data $x$, for example, when calculating the distance between generated and observed data $d(x_o, x)$. In practice, however, this is usually not the case, because the data is of high dimensionality and computationally costly to process. Therefore, summary statistics $s(x)$ are used in place of the full data $x$. 

In particular cases, a summary statistic $s(x)$ can be chosen for the data $x$ such that no information is lost. Then $s(x)$ is called a sufficient statistic. More formally, a summary statistic $s$ is sufficient for the model parameters $\theta$ if the conditional distribution of the parameters $\theta$ given the statistics $s$ does not depend on the data $x$: 
\begin{align}
p(\theta | s(x), x) &= p(\theta | s(x)). 
\end{align}
If this condition holds, it is possible to use summary statistic in place of the actual data with no effect on the performance of the ABC algorithm. In practice, it is often difficult to find sufficient statistics as data and models are usually highly complex. A review on the selection of summary statistics in the context of ABC is given by \cite{blum2013}. 

If statistics are not sufficient this adds another layer of approximation to the estimation of the posterior. For the estimation of the model evidence this has important implications as it has been shown that comparing two models based on insufficient summary statistics can be problematic \citep{didelot2011, robert2011}. Given a summary statistic $s$, data $x$ and two competing models $\mathcal{M}_1$ and $\mathcal{M}_2$ the Bayes factor can be written as \citep{didelot2011}: 
\begin{align}
B_{1, 2} &= \frac{p(x | \mathcal{M}_1)}{p(x | \mathcal{M}_2)} = \frac{p(x, s(x)| \mathcal{M}_1) p(s(x) | \mathcal{M}_1)}{p(x, s(x)| \mathcal{M}_2) p(s(x) | \mathcal{M}_2)} = \frac{p(x, s(x)| \mathcal{M}_1) }{p(x, s(x)| \mathcal{M}_2)} B_{1, 2}^s, 
\end{align}
where $B_{1, 2}^s$ is the Bayes factor based on the summary statistic. Then it becomes clear that comparing two models with the Bayes factor based on the summary statistic corresponds to the actual Bayes factor comparison if and only if $p(x, s(x)| \mathcal{M}_1) = p(x, s(x)| \mathcal{M}_2)$. Importantly, it was shown that this condition is not guaranteed even if the statistic $s$ is sufficient for both $\mathcal{M}_1$ and $\mathcal{M}_2$ \citep{grelaud2009}. 

This appears as a serious problem for model comparison in ABC in general, as pointed out by \citet{robert2011}. However, the authors further remark that although the existing methods are usually applied to problems with insufficient statistics, they seem to select the correct models. This indicates that the approach is not untrustworthy per se, but that it requires further research on methods to approximate the posterior probabilities of models \citep{robert2011}. 

\section{Summary}
I presented an introduction to Bayesian model comparison in the context of ABC. We have seen that the approaches to estimate the model posterior probabilities are based on rejection sampling methods that come with multiple limitations. Some of those limitations have been overcome by approaches using posterior density estimation in the context of learning the posterior over parameters of a model and it seems promising to extend these approaches to the model comparison problem, as it is the subject of this thesis. The next chapter introduces this extension. 

\chapter{Bayesian model comparison in approximate Bayesian computation}
\label{ch:general_methods}
This chapter presents the technical details of the approach for Bayesian model comparison in approximate Bayesian computation introduced above. The goal of the method is to approximate the posterior distribution over models given the observed data using density estimation. I will begin with formulating this objective as an optimization problem. With this formulation it will be possible to approximate not only the posterior over models, but also the posterior over the parameters of an individual model. Therefore, the method will cover the complete model selection process: it approximates the posterior over models given observed data in order to select a model; and approximates the posterior over the parameters of the selected model. 

In the second section of the chapter I will introduce methods for validating the approximated posteriors. 

\section{The optimization problem}
As explained in detail in section \ref{sec:background_bmc} the ultimate goal of the model comparison setup is to obtain a posterior over models given the observed data
\begin{align}
\label{eq:model_comparison_bayes}
p(m | x) = \frac{p(x| m) p(m)}{p(x)}, 
\end{align}
where $m$ is a discrete random variable over the model indices. Note that I change notation from $\mathcal{M}$ to $m$ and from $\mathcal{D}$ to $x$. In the ABC setting the likelihood $p(x | m)$ is unknown. However, given a prior distribution over models $p(m)$ and a simulator to generate data $x$ for every model, it is possible to draw samples $(m, x)$.  In previous approaches, this was used to estimate the model index using rejection sampling (see \ref{sec:rejection_sampling} for details). In contrast to that, I here extend the approach suggested by \citep{papamakarios2016} and learn a parametric model of the posterior over models. 

First, note that the process of generating data from the models involves the parameters of the models as well: When generating data $x$ from a model $m$, model parameters $\theta$ are sampled from a prior over parameters $p(\theta | m)$. Thus, the joint density of models, parameters and data can be factorized as
\begin{align}
\label{eq:factorization}
p(m, \theta, x) &= p(x) \; p(m | x) \; p(\theta | x, m).
\end{align}
Apart from the marginal density of the data, $p(x)$, this factorization contains the posterior over model indices, $p(m | x)$, and the the posterior over model parameters, $p(\theta | x, m)$. Now the objective is to approximate these two posteriors with regression models $q_{\psi}(m|x)$ and $q_{\phi}(\theta | x, m)$, respectively, using a set of samples $(m, \theta, x)$ to optimize the parameters $\psi$ and $\phi$. To this end, I choose as a criterion the minimization of the Kullback-Leibler divergence ($D_{KL}$) between the joint distribution $p(m, \theta, x)$ and the joint data model $p(x) q_{\psi}(m|x) q_{\phi}(\theta | x, m)$. The $D_{KL}$ between two continuous probability distributions is defined as 
\begin{align}
D_{KL}(p(x) || q(x)) &= \int p(x) \log \left(\frac{p(x)}{q(x)} \right)dx. 
\end{align}
Therefore, we have
\begin{align}
D_{KL} \Big(p(x) \; p(m | x) \; p(\theta | x, m) \;||\; p(x) q_{\psi}(m|x) q_{\phi}(\theta| x, m)\Big) &= \nonumber \\
\int p(x) \; p(m | x) \; p(\theta | x, m)  log &\left(\frac{p(m | x) \; p(\theta | x, m)}{q_{\psi}(m|x) q_{\phi}(\theta| x, m)} \right)dx \nonumber \\
= \underbrace{\int p(x) \, p(m | x) \, p(\theta | x, m) \; log \Big(p(m | x) \, p(\theta | x, m)\Big) dx}_{constant \; w.r.t \; \psi,\; \phi} - \nonumber \\ 
\int p(x) \, p(m | x) \, p(\theta | x, m) \; log &\Big(q_{\psi}(m|x) q_{\phi}(\theta| x, m) \Big)dx \nonumber \\
= - \Big\langle log \Big(q_{\psi}(m|x) q_{\phi}(\theta| x, m)\Big) &\Big\rangle_{p(x) \, p(m | x) \, p(\theta | x, m)} + const. \nonumber \\ 
= - \Big\langle log \; q_{\psi}(m|x)\Big\rangle - \Big\langle log \; q_{\phi}(&\theta| x, m) \Big\rangle + const.
\end{align}
Thus, minimizing the above $D_{KL}$ is equivalent to maximizing the expectations of the log-probability of $q_{\psi}(m| x)$ and $q_{\phi}(\theta| x, m)$ under the joint distribution $p(m, \theta, x)$ with respect to $\psi$ and $\phi$, respectively. Under the assumption that there are infinitely many samples, $N \rightarrow \infty$, and that the models $q_{\psi}$ and $q_{\phi}$ are flexible enough to approximate the posterior, minimizing the $D_{KL}$ is equivalent to minimizing the loss 
\begin{align}
L(\psi, \phi) = -\frac{1}{N} \sum_{n=1}^N log \; q_{\psi}(m_n | x_n) -\frac{1}{N} \sum_{n=1}^N log \; q_{\phi}(\theta_n | x_n, m_n), 
\end{align}
where the subscript $n$ refers to individual samples of model indices, parameters and data. Here, it is important to note that the two terms in the loss function do not depend on each other. Therefore, they can be separated into two loss functions $L(\psi)$ and $L(\phi)$ and it is possible to optimize the two models separately. Then, minimizing $L(\psi)$ will approximate $p(m | x)$ with $q_{\psi}$ and minimizing $L(\phi)$ will approximate $p(\theta | x, m)$ with $q_{\phi}$. 

\subsection{Approximation of the posterior over models}
The posterior over models $p(m | x)$ can be approximated with the model $q_{\psi}(m_n | x_n)$ by minimizing the loss $L(\psi)$. It is a distribution over a discrete model index variable and can be modeled as a multinomial distribution. I therefore use a mixture-density network (MDN) \citep{bishop1994} $q_{\psi}(m | x)$ parametrized by $\psi$, to approximate a multinomial distribution. In other words, the approximation of the model index posterior is treated as a multi-class classification problem solved with softmax regression by minimizing the cross-entropy loss 
\begin{align}
L(\psi) &= -\frac{1}{N} \sum_{n=1}^N log \; q_{\psi}(m_n | x_n) \\
&= -\frac{1}{N} \left [ \sum_{n=1}^N m_n log \; q_{\psi}(m_n | x_n) + (1 - m_n) log\;(1 - q_{\psi}(m_n | x_n)) \right].
\end{align}
Here, $m_n$ is the target model index and $q_{\psi}(m_n | x_n)$ is the posterior probability predicted by the MDN. 

The MDN takes as input the data $x$ or the summary statistics of the data $s(x)$. The output $q_{\psi}(m | x)$ is computed by a feedforward neural network $f_{\textbf{W}, \textbf{b}}(s(x))$ with weights $\textbf{W}$ and biases $\textbf{b}$. 
The layers are fully connected and the units have $tanh$ activation functions. In particular, if the activation of the last hidden layer is 
\begin{align} 
\mathbf{y} &= f_{\textbf{W}, \textbf{b}}(s(x)),
\end{align}
then the predicted posterior probability is calculated from 
\begin{align}
q_{\psi}(m | x) &= \text{softmax}(\textbf{W}_{out}\mathbf{y} + \textbf{b}_{out}).
\end{align}
where the subscript \textit{out} refers to the weights and biases of the output layer. 

In summary, to train the MDN $q_{\psi}(m | x)$ for approximating the posterior over models $p(m | x)$ we use samples $(x_n, m_n)$ to minimize the cross entropy loss with respect to the parameters $\psi = (\mathbf{W}, \mathbf{b}, \mathbf{W}_{out}, \mathbf{b}_{out})$. 
\subsection{Approximation of the posterior over model parameters}
The posterior over model parameters given the observed data and the model index, $p(\theta | x, m)$, is a continuous probability distribution that can take any form, dependent on the model class. Therefore, the model $q_{\phi}(\theta| x, m)$ that is used to approximate it is defined as a MDN of a mixture of Gaussians, 
\begin{align}
q_{\phi}(\theta| x, m) &= \sum_k \; \alpha_k \; \mathcal{N}(\mathbf{\theta} | \mathbf{m}_k, \mathbf{S}_k). 
\end{align}
The mixing coefficients $\mathbf{\alpha}_k$, the means ${\mathbf{m}_k}$ and the covariances ${\mathbf{S}_k}$ are computed by a feedforward network $f_{\mathbf{W}, \mathbf{b}}(s(x), m)$ with multiple hidden layers summarized by weights $\textbf{W}$ and biases $\textbf{b}$. This network takes the summary statistics of the data and the model index as input features. 
The layers are fully connected and the units have $tanh$ activation functions. 

The outputs $\mathbf{\alpha}_k$, ${\mathbf{m}_k}$ and ${\mathbf{S}_k}$ are computed in an approach as used by \cite{papamakarios2016}: If $\mathbf{y} = f_{\textbf{W}, \textbf{b}}(s(x),  m)$ is the activation of the last hidden layer, then the mixing coefficients are given by
\begin{align}
\mathbf{\alpha}_k &= \text{softmax}(\mathbf{W}_{\mathbf{\alpha}_k} \mathbf{y} + \mathbf{b}_{\mathbf{\alpha}_k})
\end{align}
where the softmax activation function ensures that the mixture weights sum up to 1. The means $\mathbf{m}_k$ are calculated as a linear combination of $\mathbf{y}$:
\begin{align}
\mathbf{m}_k &= \mathbf{W}_{\mathbf{m}_k} \mathbf{y} + \mathbf{b}_{\mathbf{m}_k}. 
\end{align}
The covariances of the mixture-density need to be symmetric and positive definite. To ensure these properties, the network does not output the covariances directly, but the Cholesky transform of their inverses, 
\begin{align}
\mathbf{S}_k^{-1} = \mathbf{U}_k^T\mathbf{U}_k. 
\end{align}
In particular, the activation $\mathbf{y}$ of the last hidden layer is transformed and arranged in an upper triangular matrix as follows. From the last layer activation there is another linear readout to the matrix elements, and additionally, the diagonal elements are passed through an exponential function to ensure positivity:
\begin{align}
\mathbf{u}_k^{diag} &= \exp(\mathbf{W}_{\mathbf{u}_k^{diag}} \mathbf{y} + \mathbf{b}_{\mathbf{u}_k^{diag}}) \\ 
\mathbf{u}_k^{utri} &= \mathbf{W}_{\mathbf{u}_k^{utri}} \mathbf{y} + \mathbf{b}_{\mathbf{u}_k^{utri}}. 
\end{align}
Then the Cholesky transform matrix $\mathbf{U}_k$ is set up by assigning $\mathbf{u}_k^{diag}$ to the diagonal elements and $\mathbf{u}_k^{utri}$ to the upper triangular elements, leaving the remaining elements of the matrix zero. 

In summary, the posterior over the model parameters, $p(\theta | x, m)$, is approximated by the MDN $q_{\phi}(\theta | x, m)$ parametrized by a mixture of Gaussians. It is trained by maximizing the average log probability of the model, which is equivalent to minimizing the average loss 
\begin{align}
L(\phi) &= -\frac{1}{N} \sum_{n=1}^N log \; q_{\phi}(\theta| x, m) \\ 
&= -\frac{1}{N} \sum_{n=1}^N log \sum_k \; \alpha_k \; \mathcal{N}(\mathbf{\theta} | \mathbf{m}_k, \mathbf{S}_k)
\end{align}
with respect to the parameters  
\begin{align*}
\phi = (\mathbf{W}, \mathbf{b}, \mathbf{W}_{\mathbf{\alpha}}, \mathbf{b}_{\mathbf{\alpha}}, \mathbf{W}_{\mathbf{m}_k}, \mathbf{b}_{\mathbf{m}_k}, \mathbf{W}_{\mathbf{\mathbf{u}_k^{diag}}}, \mathbf{b}_{\mathbf{\mathbf{u}_k^{diag}}}, \mathbf{W}_{\mathbf{\mathbf{u}_k^{utri}}}, \mathbf{b}_{\mathbf{\mathbf{u}_k^{utri}}}).
\end{align*} 

\subsection{Implementation}
The approximation of the posterior over models and the posterior over parameters using MDNs is implemented in Pytorch \citep{pytorch}. The loss functions of the MDNs are minimized by backpropagation of the gradients and adapting the parameters with stochastic batch optimization using Adam \citep{adam}. The batch size, number of epochs and, in case of the mixture of Gaussians MDN, also number of mixture components are chosen heuristically. The complete code of the project is publicly available on GitHub (\url{https://github.com/janfb/mcabc}).

Given a prior over the models to be compared, a prior over parameters for every model and a simulator to generate data from parameters, the training data is generated from the models in the following way: A model index $m_n$ is sampled from the prior over models, prior parameters $\theta_n$ are sampled from the corresponding prior over parameters and data $x_n$ is generated data from the model. This gives a single joint sample $(m_n, \theta_n, x_n)$. If the MDN takes summary statistics as input, the summary statistics $s(x_n)$ are calculated for every data point $x_n$. 

To improve the convergence of the loss function the training data $x$ is $z$-transformed. Similarly, in case of the approximation of the posteriors over model parameters, the parameters $\theta_n$ are $z$-transformed to ease the initialization of the network weights and to improve the convergence of the training procedure. After the training, the approximated posterior is transformed back into the original parameter range. When predicting unseen test data, the test data is $z$-transformed using the mean and standard deviations of the training data set. 

\section{Posterior validation}
\label{sec:posterior_validation}
The above method aims at approximating the true posterior over model indices and, for every model, the posterior over model parameters. I derived a loss function that will lead to an exact approximation in case of an infinite amount of data. As this does not hold in practice, the approximated posteriors have to be validated. Here, I distinguish between two scenarios. First, the method can be validated on an example problem for which the ground-truth posteriors are available. This will demonstrate that the method works in a tractable case. Second, there are validation steps for the use-case, i.e., when there are no ground-truth posterior available. 

In either case, for the validation process a large number of model indices from a model prior and parameters from the corresponding priors are sampled to generate a test data set $\{m_n, \theta_n, x_n\}_{n=1}^N$. 

\subsection{Validation with ground-truth posterior}
With a ground-truth posterior at hand the approximated posterior can be validated directly. The methods of validation differ for the discrete model index posterior and the continuous model parameter posterior. 

\subsubsection{Posterior over models} To validate the approximated model index posterior I calculate $q_{\psi}(m_n | x_n)$ for every test sample $(m_n, \theta_n, x_n)$ and compare it to the exact posterior probability $p(m_n | x_n)$. The exact posterior probability is calculated as  
\begin{align}
\label{eq:model_index_posterior}
p(m_n | x_n) &= \frac{p(x_n | m_n) p(m_n)}{\sum_{i=1}^M p(x_n | m_i) p(m_i)}, 
\end{align}
where $p(x_n | m_n)$ is the evidence under the model $m_n$ and $p(m_n)$ is the prior probability. The sum in the denominator runs over all model indices. 

\subsubsection{Posterior over model parameters} The continuous model parameter posterior is approximated with a mixture of Gaussians. For each test sample $x_n$ we obtain the predicted posterior $q_{\phi}(\theta | x_n)$ and the exact posterior. As a first validation we compare the means and the variances. If the posterior is multidimensional, means and variances are compared for each marginal separately. Additionally, we calculate the Kullback-Leibler divergence ($D_{KL}$) between the exact posterior and the estimated posterior, normalized by the $D_{KL}$ between the exact posterior and the prior. This normalization improves the interpretability of this validation step: The $D_{KL}$ between the posterior and the prior is a baseline measure for how well one would do without having seen any data. 

In some cases the $D_{KL}$ is not tractable, even if the ground-truth posterior is available. Then it is approximated as the expectation over the logarithm of the two probability density functions:
\begin{align}
D_{KL}(p(x) || q(x)) &= \mathbb{E} \left[\log \left(\frac{p(x)}{q(x)} \right) \right]_{p(x)} \\
&\approx \frac{1}{N} \sum_{i=1}^N \log \left(\frac{p(x_i)}{q(x_i)} \right).
\end{align}

In case of a multidimensional posterior, it is sensible to not only validate the means and variances of the marginals, but also the overall covariance structure. To this end, I investigate the marginals defined by the direction of the eigenvectors of the covariance matrix, $\nu_{max}$ and $\nu_{min}$, corresponding to the largest and smallest eigenvalues, $\lambda_{max}$ and $\lambda_{min}$. To test the accuracy of the approximation of the posterior one can compare the sample distributions along the marginals. If the exact posterior covariance matrix is estimated accurately then the distributions of samples, from the exact and from the estimated posterior, projected along the eigenvector marginals of the estimated posterior should be very similar. Therefore, I show how samples from both posteriors are distributed along these two marginals and compare again the sample means and variances of the projections.  Furthermore, I quantify how much variance the estimated eigenvector marginals capture of the exact samples by comparing the variance of the projected samples with the corresponding eigenvalue of the exact posterior. 
\subsection{Validation without ground-truth posterior}
In the use-case scenario there is no ground-truth posterior available. Still, there exist several methods to test the quality of the estimation. 

\subsubsection{Posterior over models} For the discrete posterior over model indices I examine how the estimated posterior probability depends on the prior. Multiple test data sets, each with a large number of test samples $(m_n, \theta_n, x_n)$, are generated using a range of different prior probabilities. If the posterior is approximated correctly, the model prior probabilities are reflected in the relative frequencies of the predicted model classes. To give an example, if there are models $A$ and $B$ and the prior is $p(A)=0.1$, we expect that the method predicts a large posterior probability for model $A$ for approximately $10\%$  of the test samples. 

\subsubsection{Posterior over model parameters} Regarding the posterior over model parameters I consider two methods of validation. 

The first one only applies to one-dimensional posteriors (or marginals) and is based on the following observation: If one samples from a random variable and then calculates the quantiles of the samples with respect to its probability distribution, then the quantile distribution follows a uniform distribution \citep{cook2006}. Additionally, consider a parameter that is sampled from a prior distribution, the corresponding data that is generated from a model and the exact posterior over parameters calculated given the data. Then the parameter that was initially sampled from the prior is also a sample from the posterior distribution. Therefore, if we consider all the test samples generated from a prior, then the quantiles of these samples under the exact posterior are distributed uniformly. Thus, testing whether the distribution of quantiles of the prior parameters under the estimated posterior is uniform is a sanity check for the estimated posterior. This does not generalize to multiple dimensions, however, in that case one can perform the quantile check for the one-dimensional marginal over each model parameter separately.

The second validation step investigates the credible intervals of the estimated posterior. A credible interval of a distribution of a random variable is defined by a subjective probability that corresponds to an interval into which the random variable falls with that probability. For each parameter $\theta_n$ from the test data set I consider whether it falls in a credible interval of the estimated posterior for a set of different subjective probabilities, e.g., $0.05, 0.1, \ldots, 0.95$. If the estimated posterior is close to the true posterior, the relative frequency with which the sampled parameter falls into a credible interval should correspond to the subjective probability of that interval. For example, if we generate 100 test samples $\theta_n$, predict the posterior given data $x_n$ for each of them and check whether the test samples fall into the $90\%$ credible interval of the posterior, this should happen approximately 90 ($90\%$) times. 

The concept of credible intervals can be extended to credible regions in multiple dimensions. However, we will apply it separately to the one-dimensional marginals in case of a multidimensional posterior. \\[0.5mm]

This concludes the presentation of the methods used in the new approach. Two separate loss functions have been derived, one that optimizes an MDN for the posterior over models, and another one optimizing an MDN for the posterior over model parameters. The approximated posteriors can be validated with the validation methods presented above. In the following two chapters the performance of this optimization will be demonstrated on two examples, a theoretical example with ground-truth posterior and a realistic use-case scenario without ground-truth posterior. 

\chapter{Model comparison on a tractable example}
\label{ch:toy_example}
To explore the performance of the model comparison method I choose an example problem for which the ground-truth is known. This example is the comparison between a Poisson model and a negative binomial (NB) model. Both models are often used to model the occurrence of events in given time intervals. The Poisson distribution has by definition equal mean and variance, whereas the NB model is overdispersed, i.e. its variance is larger than the mean. This is why the NB model is often used as an alternative to the Poisson model in cases where the sample variance is expected to be larger than the mean, e.g. when modeling the spike count variability in neural spiking data. 

The goal of this example problem is to demonstrate that with the proposed method it is possible to compare two competing models given observed data, to select the one that is more likely based on the posterior probability, and to find the posterior over parameters for this model. Because the central difference between the two models is the overdispersion of the NB model, the difficulty of the comparison can be scaled by scaling the amount of overdispersion. How this is done is illustrated in the following section after presenting the two models in more detail. 

\section{Methods}
\subsection{Poisson model}
The Poisson model describes the data with a Poisson random variable $X$, that expresses the probability of the number of events occurring in a certain time interval. The events are assumed to occur independently and with a constant predefined rate. Given a rate $\lambda$ and a fixed time interval $T$ the probability of $x$ events to occur in $T$ is given by the probability mass function 
\begin{align}
\label{eq:poisson}
p(X=x | \lambda) &= \frac{\lambda^{x} e^{-\lambda}}{x!}.
\end{align}
The mean and the variance of the Poisson distribution are given by $\lambda$. When observing multiple independent trials the likelihood of a series of observations $X=x_i$ is given by
\begin{align}
\label{eq:poisson_likelihood}
p(X | \lambda) &= \prod_{i=1}^N \frac{\lambda^{x_i} e^{-\lambda}}{x_i!} = \frac{\lambda^{(\sum x_i)} e^{-N\lambda}}{\prod_i x_i!}
\end{align}
The Poisson distribution belongs to the exponential family. Therefore it has a conjugate prior and a closed form posterior. As a prior over the Poisson rate $\lambda$ I choose the conjugate Gamma prior 
\begin{align}
\label{eq:poisson_prior}
p(\lambda | k, \theta) &= \frac{1}{\Gamma (k)\theta^k} \lambda^k e^{-\frac{\lambda}{^\theta}}
\end{align}
with shape parameter $k$ and scale parameter $\theta$. For the purpose of model comparison the evidence of the Poisson model under the observed data is of special interest. The evidence is defined as the product of likelihood and prior integrated over all prior parameters
\begin{align}
p(x | k, \theta) &= \int_0^{\infty} p(x | \lambda) p(\lambda | k, \theta) d\lambda \\
&= \frac{1}{\Gamma(k) \theta^k} \frac{1}{\prod x_i!} \int_0^{\infty} \lambda^{k + \sum x_i - 1}e^{-\lambda(N + \theta^{-1})} d\lambda \\
&= \frac{\Gamma(k + \Sigma x_i)}{\Gamma(k) \theta^k} \frac{(N + \theta^{-1})^{-(k + \Sigma x_i)} }{\prod x_i!}
\end{align}
and it depends on the choice of hyper parameters of the Gamma prior, $k$ and $\theta$. 

Following Bayes rule, the posterior distribution of the Poisson rate $\lambda$ given the observed data $X$ is the product of likelihood and prior normalized by the evidence. Because I chose a conjugate Gamma prior the posterior follows a Gamma distribution: 
\begin{align}
p(\lambda | X, k, \theta) &= \frac{p(x|\lambda, k, \theta) p(\lambda | k, \theta)}{p(x | k, \theta)} \\ 
&= \frac{\frac{\lambda^{k + \sum x_i - 1}}{\Gamma(k) \theta^k} \frac{e^{-\lambda(N + \theta^{-1})}}{\prod x_i!}}{\frac{\Gamma(k + \Sigma x_i)}{\Gamma(k) \theta^k} \frac{(N + \theta^{-1})^{-(k + \Sigma x_i)} }{\prod x_i!}} \\ 
&= \frac{\lambda^{k + \sum x_i - 1} e^{-\lambda(N + \theta^{-1})}}{\Gamma(k + \Sigma x_i) (N + \theta^{-1})^{-(k + \Sigma x_i)}} \\ 
&= Gamma(k + \Sigma x_i, (N + \theta^{-1})^{-1}).
\end{align}
\subsection{Negative binomial model}
\label{sec:binom_model}
The negative binomial model (NB) describes the data with a negative binomial random variable. A negative binomial random variable gives the number of successes in a sequence of independent and identically distributed Bernoulli trials before a specified number of $r$ failures occurs. To give an example, in a coin toss experiment, the number of times tails comes up before three heads have occurred follows a negative binomial distribution with $r=3$. For success probability $p$ and a number of failures $r$ the probability to observe x successes is given by 
\begin{align}
\label{eq:nb}
p(x | r, p) &= \binom{x + r - 1}{x} (1 - p)^r p^x
\end{align}
and for a sequence of independent trials $X$ it is given by
\begin{align}
\label{eq:nb_likelihood}
p(X | r, p) &= (1 - p)^{Nr} p^{\Sigma x_i} \prod_{i=1}^N \binom{x_i + r - 1}{x_i},
\end{align}
defining the likelihood. The mean is given by $\frac{pr}{1-p}$ and the variance by $\frac{pr}{(1-p)^2}$. 

Regarding the choice of the prior I note that the NB is in the exponential family as well if $r$ is fixed, with a conjugate Beta prior on $p$. However, by setting $r$ to a fixed value the additional degree of freedom on the variance, and thereby the crucial difference to the Poisson model, is lost. Because the core idea of this example problem is to perform a model comparison based on the overdispersion present in the data I choose a prior for both $p$ and for $r$. 

To this end, I formulate the NB model differently. A negative binomial random variable can also be described as a Poisson random variable for which the rate $\lambda$ is itself a Gamma random variable with shape parameter $k=r$ and scale parameter $\theta=\frac{p}{1-p}$: 
\begin{align}
p_{NB}(X=x | r, p) &= \int_{\lambda} p_{Poisson}(x|\lambda) \; p_{Gamma}(\lambda | k=r, \theta=\frac{p}{1-p}) d\lambda.
\end{align}
To have a similar setup as in the Poisson case and to have more flexibility in choosing the prior parameters I express the NB model as this mixture of a Poisson with a Gamma rate and use again Gamma priors on the shape and scale parameter of Gamma: 
\begin{align}
\label{eq:nb_prior}
p_{NB}(x | k, \theta) &= p_{Poisson}(x | \lambda)p_{Gamma}(\lambda | k, \theta) \\ 
\text{with} \;\; k &\sim Gamma(k_2, \theta_2) \\
\text{and} \;\; \theta &\sim Gamma(k_3, \theta_3)
\end{align}
With this formulation the evidence integral of the NB model is calculated over three variables, the two prior variables $k$ and $\theta$ and the Poisson rate $\lambda$ of the Poisson-Gamma mixture. It can be reduced to two integrals by going back to the original formulation in terms of the prior variables $p$ and $r$ by a change of variables from $k$ to $r$ and from $\theta$ to $p$:
\begin{align}
\label{eq:nb_evidence}
p(x) &=  \int_k \int_{\theta} \int_{\lambda} \; p_{Poisson}(x|\lambda) \; p_{Gamma}(\lambda | k, \theta)  \; p(k) \; p(\theta) d\lambda d\theta dk  \\
&= \int_k \int_{\theta} p_{NB}(x| r, p)  \; f_k(k) \;  f_{\theta}(\theta) \; d\theta dk \\ 
&= \int_r \int_p p_{NB}(x|r, p) \; \frac{1}{(1-p)^2} \;  f_{\theta}\left(\frac{p}{1-p}\right)  \; f_k(r) \; drdp,
\end{align}
where $r$ is a defined as a real number. 

There is no closed form solution for this integral. And in general, with this alternative formulation of the NB model there is no closed form posterior distribution. Therefore, to calculate the evidence of the NB model I calculate the above integral numerically. And accordingly, the exact joint posterior over $k$ and $\theta$ is also determined numerically. To do so for a single pair of $k$ and $\theta$, I again take the old formulation of the NB model for the calculation of the likelihood in order to avoid the additional integral:
\begin{align}
\label{eq:nb_posterior}
p(k, \theta | x) &= \frac{p_{NB}(x | k, \theta) p(k) \; p(\theta)}{p(x)} \\ 
&= \frac{p_{NB}(x| r, p)  \; p(k) \; p(\theta)}{p(x)}, 
\end{align}
with $r=k$ and $p=\frac{\theta}{1 + \theta}$. To obtain an overall representation of the exact NB model posterior, the posterior probabilities for all $(k, \theta)$ pairs are calculated on a large grid of values almost over the complete range of the prior distributions. The corresponding cumulative density function is approximated numerically by taking the cumulative sum over the grid and normalizing by the total sum. Samples are generated using inverse transform sampling: the marginal density $p(k)$ is obtained by numerically integrating the grid of PDF values over $\theta$ and the conditional density $p(\theta | k)$ is obtained by calculating $p(\theta | k) = \frac{p(k, \theta)}{p(k)}$ for every $k$. Then, pairs of samples $(k_i, \theta_i)$ are generated by sampling $k_i$ from $p(k)$ and $\theta_i$ from $p(\theta | k_i)$ using inverse transform sampling via the approximated CDF array. 

\subsection{Controlling the difficulty of model comparison}
I apply the model comparison method to a Poisson model and a Negative Binomial (NB) model. The main difference between the two models is the overdispersion in the NB model. Accordingly, the difficulty of the problem can be controlled by changing the amount of overdispersion: the more overdispersed the NB model the easier it will be distinguishable from the Poisson model. However, scaling up the variance in the NB model will result in a higher mean as well. In order to base the comparison mainly on overdispersion and not on mean differences in the data, the sample means of the data sets generated from the two models should be equal on average. 

The average sample mean and sample variance of the models are controlled by the hyper-parameters of the priors. There is a Gamma prior on the Poisson rate and there are Gamma priors on the Gamma shape and scale in the Poisson-Gamma mixture of the NB model: 
\begin{align}
\lambda &\sim Gamma(k_1, \theta_1) \\
r = k &\sim Gamma(k_2, \theta_2) \\
\frac{p}{1-p} = \theta &\sim Gamma(k_3, \theta_3).
\end{align}
Starting from the constraint that the expectation over the NB mean equals the expectation over the Poisson mean I derive a condition on the prior hyper-parameters: 
\begin{align}
\mathbf{E}[\mu_{NB}] &= \mathbf{E}[\mu_{Poi}] \\
\mathbf{E}[r \frac{p}{1-p}] &= \mathbf{E}[\lambda] \\
\mathbf{E}[r] \mathbf{E}[\theta] &= k_1 \theta_1 \\ 
k_2 \theta_2 k_3 \theta_3 &= k_1 \theta_1.
\end{align}
Similarly, I can find the hyper-parameters that have to be changed in order to scale up the expected variance in the NB model. With $r=k$, $p=\frac{\theta}{\theta + 1}$ and $\theta = \frac{p}{1-p}$, the variance can be written in terms of $k$ and $\theta$: 
\begin{align}
\sigma_{NB}^2 &= \frac{pr}{(1-p)^2} = r \left(\frac{p}{1-p}\right)^2 \frac{1}{p}\\
&= k \theta^2 \frac{\theta + 1}{\theta} = k (\theta^2 + \theta).
\end{align}
Then we have for the expectation 
\begin{align}
\mathbb{E}[\sigma_{NB}^2] &= \mathbb{E}[k (\theta^2 + \theta)] \\
&= \mathbb{E}[k] \; \mathbb{E}[\theta^2 + \theta] \\ 
&= k_2 \theta_2 \mathbb{E}[\theta^2 + \theta]. 
\end{align}
Thus, I can choose to set $\theta_2$, $k_3$ and $\theta_3$ heuristically, scale the expected variance as a function of $k_2$ and adapt $k_1$ accordingly to have equal expected means: 
\begin{align}
\mathbb{E}[\sigma_{NB}^2] &= f(k_2 | \theta_2, k_3, \theta_3) \\ 
k_1 &= \frac{k_2 \theta_2 k_3 \theta_3}{\theta_1 }. 
\end{align}
This enables me to scale the model comparison problem from a relatively easy example (large $k_2$) to a relatively difficult one (small $k_2$). 

Overall, I have defined a tractable example problem for which we know all the quantities needed to perform model comparison analytically or numerically exact. 

\subsection{Data generation}
To generate data from the models, prior distributions over the model parameters have to be defined. The Poisson model comes with a single Gamma prior with shape $k_1$ and scale $\theta_1$. The NB model comes with two Gamma priors, one for the internal Gamma shape, with shape $k_2$ and scale $\theta_2$ and one for the internal Gamma scale, with shape $k_3$ and scale $\theta_3$. Above, we have seen how to choose the prior hyper-parameters to obtain different amounts of overdispersion in the NB model while keeping the expected sample mean in both models equal. For the model comparison example setup I now generate data for two scenarios: a relatively simple case with $k_2=20$ such that there is large dispersion in the NB model and a relatively difficult case with $k_2=1$ such that the overdispersion is low and the two models are quite similar. With $k_2$ fixed and the remaining prior parameters set accordingly, prior parameters are sampled from the priors and count data is sampled from the models. 

The count data can be sampled from the models with different sample sizes. For a given set of sampled prior parameters I decided on a set of $C=100$ counts, such that a single data point from the model will be a vector of 100 numbers. 

\subsection{Summary statistics} The mixture-density networks (MDN) that approximate the posteriors do not take the entire data vector as input. Instead, summary statistics are calculated. Ideally, the summary statistics are sufficient such that no information is lost (see section \ref{sec:summary_stats} for details). Because the Poisson distribution is in the exponential family, there is a sufficient statistics: the sum over the sampled vector of counts. However, this does not apply to the NB model as defined here (see section \ref{sec:binom_model} above). For this reason and because the model comparison should be based on the amount of overdispersion present in the data, I take the sample mean and the sample variance as the summary statistics: 
\begin{align}
s(\mathbf{x}) &= [\bar{\mathbf{x}}, s^2],  \\ 
\bar{\mathbf{x}} &= \frac{1}{M}\sum_{i=1}^M x_i,  \\ 
s^2 &= \frac{1}{M - 1}\sum_{i=1}^M (x_i - \bar{\mathbf{x}})^2. 
\end{align}
\subsection{Training procedure}
The approximation of the posterior over models and the posterior over model parameters using MDNs is explained in detail in section \ref{ch:general_methods}. To train both MDNs, a single data set with $N=100000$ data points each holding $C=100$ counts is generated in the following way. A model index is sampled from a uniform prior and the prior parameters are sampled from the corresponding priors. With the sampled parameters data is generated from the model. This is done for both, the simple scenario ($k_2=20$) and the difficult scenario ($k_2=1$) as explained above. The summary statistics are calculated for each data point and finally, the resulting data set and the parameters are $z$-transformed. 

The MDN for approximating the posterior over models is defined with two input units for the summary statistics, a single hidden layer with 10 units, and two output units for the estimated posterior probabilities. With a learning rate of $\eta=0.01$ and a minibatch size $n_{batch}=N/100$ the weights are optimized using Adam \citep{adam} with the default parameter settings over $n_{epochs}=100$ epochs. 

Regarding the MDN for the posterior over model parameters I implement separate MDNs for each model parameter posterior: one for the one-dimensional posterior of the Poisson model, with two input units, a single hidden layer with 10 units and parameters for a univariate mixture of Gaussian with three components as output; and another one for the two-dimensional posterior of the NB model, with two input units, 2 hidden layers of 10 units each and a three-component multivariate mixture of Gaussian as output. In both cases we use a learning rate of $\eta=0.01$, $n_{batch}=N/100$,  $n_{epochs}=100$ and Adam with default parameters. 

In principle, it is possible to define a single MDN to approximate the posteriors for both models. The network would take the summary statistics and the model index as input, share the weights in the first layers and route to different output layers dependent on the given model index. The different output layers would define different mixture of Gaussians for approximating the corresponding posterior distribution. This procedure would be computationally less expensive as weights can be shared in the first layers of the network. 

\subsection{Testing procedure}
The performance of the model comparison method presented above is validated by comparing the posterior distributions approximated by the MDNs with the exact posterior distributions that are available for the example problem. A detailed explanation of the procedure of posterior validation is given in section \ref{sec:posterior_validation}. 

For the validation process, a test data set of size $N=1000$ is generated. The data is generated by sampling model indices from the model prior and model parameters from the corresponding priors to obtain data from the corresponding model, resulting in test samples $(m_n, \theta_n, x_n)$. The hyper-parameters of the priors and the parameters for the $z$-transform are the same as in the training set. 

To validate the discrete posterior over models I compare the predicted and exact posterior probabilities. The exact posterior probabilities are calculated in closed form for the Poisson model and numerically for the NB model. In comparison, the posterior probabilities are predicted with a standard sequential Monte Carlo for model comparison (SMC) \citep[section \ref{sec:rejection_sampling}]{toni2009} using the \textit{pyABC} toolbox by \citet{klinger2017pyabc}. For the prediction with these two methods the number of samples was restricted to the same number of training samples used by our method. Additionally, I investigate how the posterior probability of each model class averaged over the test data set depends on the prior over models. 

In case of the parameter posteriors there are two MDNs to be validated. The one-dimensional posterior of the Poisson model is evaluated by comparing means and variances given in closed form, and by calculating the Kullback-Leibler divergence ($D_{KL}$) between the exact and the approximated posterior. Furthermore, I evaluate the posterior quantiles and the posterior credible intervals. 

These comparisons are less straight forward in case of the two-dimensional posterior of the NB model for which the exact posterior has to be approximated numerically (described in section \ref{sec:binom_model}). Means, covariances and the $D_{KL}$ are estimated from a large set of samples and compared for each marginal. And as before, I evaluate the posterior quantiles and credible intervals for every marginal. Finally, the covariance matrix of the approximated joint posterior $q_{\phi}(k, \theta | x_n)$ is validated by comparing the largest and smallest covariance eigenvectors and eigenvalues of the exact and the approximated posterior (described in section \ref{sec:posterior_validation}). 

\section{Results}
As a tractable example I set up a model comparison between a Poisson model and a negative binomial model. Both models aim at explaining the counts of events of a discrete random variable. While the Poisson model assumes that mean and variance of the event counts are equal, the NB model assumes that the variance is larger than the mean. In the following, I will first show the data generated by the two models and results on how to change the difficulty of the example problem, before turning to the results of the model comparison. 

\subsection{Simulated data and difficulty scaling}
\begin{figure}[H]
\centering
\includegraphics[width=\textwidth]{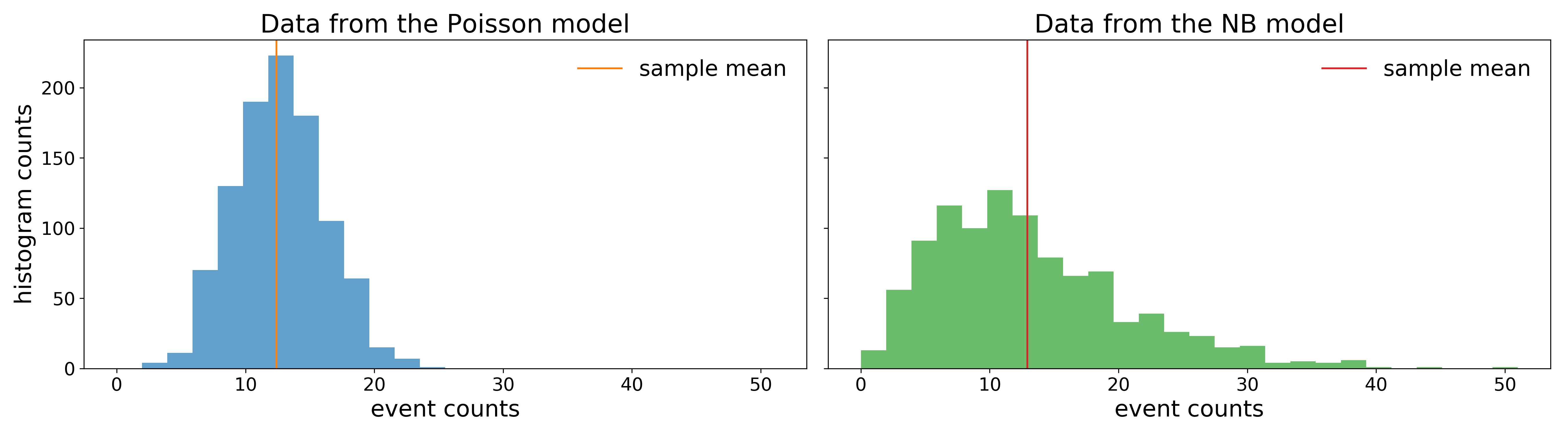}
\caption{Histograms of event counts sampled from the Poisson model and the negative binomial model.}
\label{fig:count_histrograms}
\end{figure}

Drawing a single sample of 1000 counts from the Poisson model and from the negative binomial model (NB) for a representative set of prior parameters results in the data shown in figure \ref{fig:count_histrograms}. 
The counts observed under the Poisson model have equal sample mean and variance, whereas the sample mean is smaller than the sample variance for the negative binomial model, i.e. the data is overdispersed. 

\begin{figure}[H]
\centering
\includegraphics[width=\textwidth]{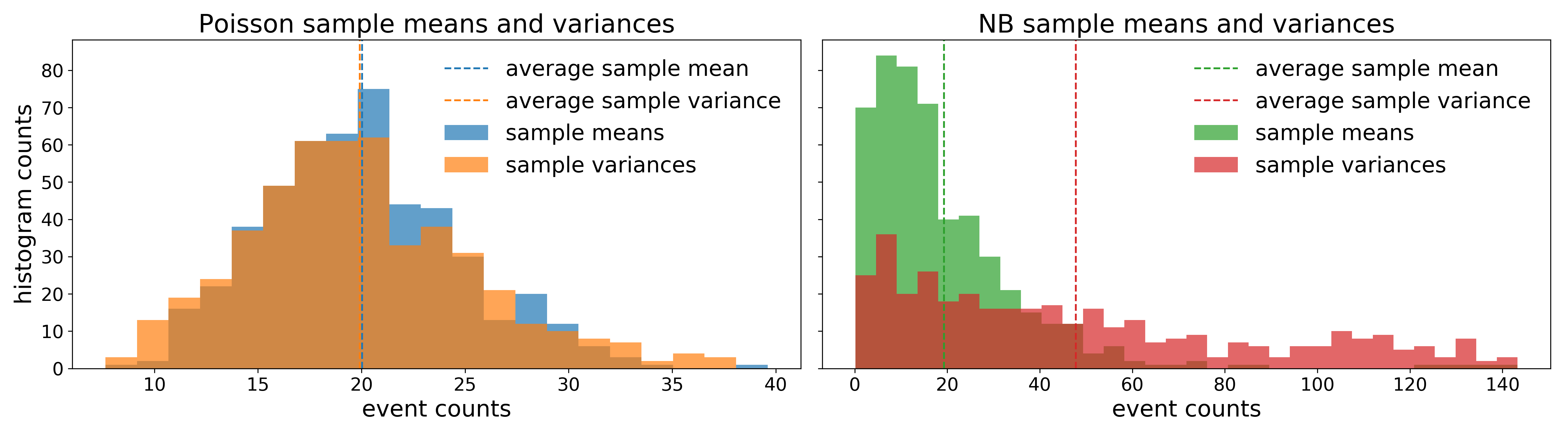}
\caption{Histograms of sample means and variances of the Poisson model and the negative binomial model.}
\label{fig:stats_histograms}
\end{figure}

This relation becomes even clearer when sampling multiple times and plotting the sample mean and variance for every sample for each model, as shown in figure \ref{fig:stats_histograms}. The distribution of sample means and variances are similar for the Poisson model, whereas for the NB model the distribution of sample variances shows a long tail towards larger values.  

\begin{figure}[H]
\centering
\includegraphics[width=\textwidth]{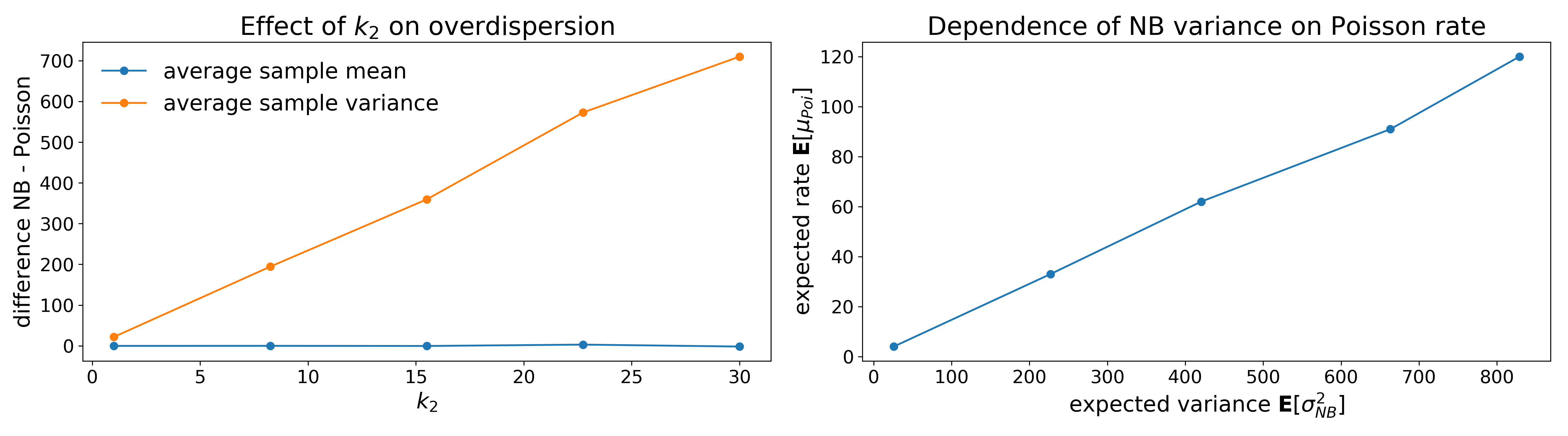}
\caption{Left, the effect of increasing $k_2$ on the difference in sample mean (blue) and variance (orange) of the two models. Right, the increasing NB sample variance plotted against the rate parameter of the Poisson model (right).}
\label{fig:difficulty_params}
\end{figure}

Thus, the amount of overdispersion in the NB model determines the difficulty of the model comparison. To control the difficulty systematically I investigated the effect of the hyper-parameters of the Gamma priors on the expected sample means and variances of the two models. By increasing the Gamma shape $k_2$ and adapting the remaining parameters, the expected sample variance of the NB model increases while the expected sample means of both models remain constant (figure \ref{fig:difficulty_params}, left). Concurrently, one can observe how the rate of the Poisson model increases with increasing $k_2$ such that the expected sample means are equal (figure \ref{fig:difficulty_params}, right).
As illustrated above, the model comparison problem becomes easier when $k_2$ increases. In the following I investigate a \textit{difficult} case, for which we set $k_2=1$ and an \textit{easy} case, for which we set $k_2=20$. First, the results for the posterior over the model indices are presented. 
\subsection{Approximating the posterior over models}
For both scenarios I trained a mixture-density network (MDN) to predict the posterior probability of a model given the data. The predictions are validated against the exact posterior probabilities that are calculated analytically for the Poisson model and numerically for the NB model. 
\begin{figure}[H]
\centering
\includegraphics[width=\textwidth]{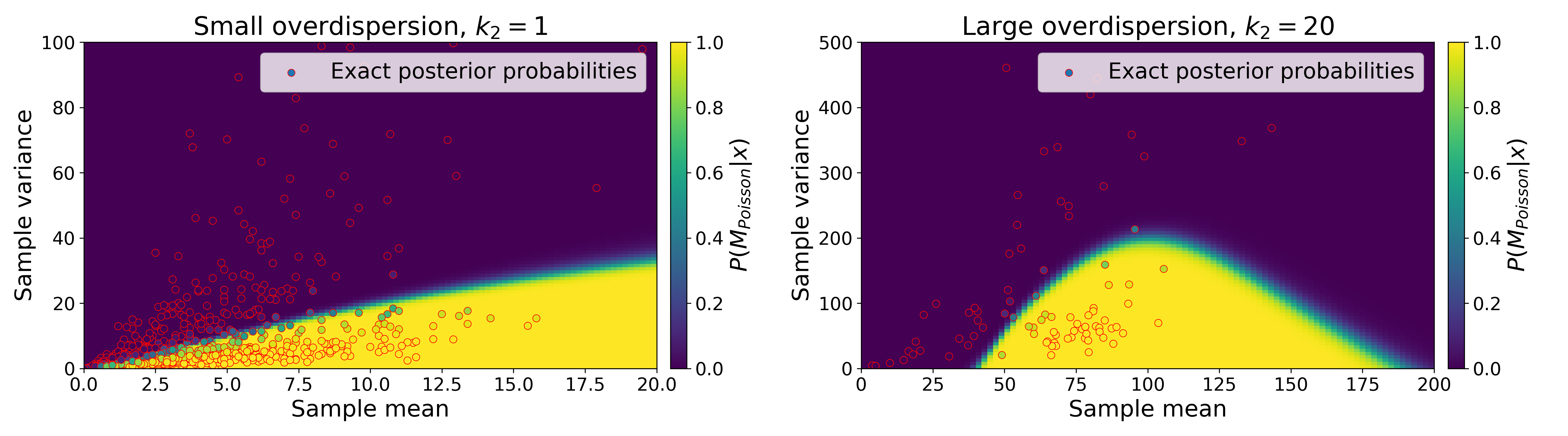}
\caption{Visualization of the mapping from summary statistics to posterior probabilities learned by the MDN with corresponding exact posterior probabilities plotted as circles with same color code.}
\label{fig:network_visualization}
\end{figure}

To give a general overview of the performance, the predicted posterior probability of the Poisson model over a representative grid of values is shown in figure \ref{fig:network_visualization}. For both scenarios there is a clear transition in the data space from predicting one model to predicting the other model. In the \textit{difficult} scenario (figure \ref{fig:network_visualization} left) the posterior probability for the Poisson model is high for data with low variance; the transition to a high probability for the NB model happens as the variance exceeds the mean, i.e. as the data becomes overdispersed. In the \textit{easy} scenario (figure \ref{fig:network_visualization} right) the ranges of sample means and variances are larger. Here, one can observe that there is a region with low variance and low mean in which the posterior probability for the NB model is high. The transition to a high posterior probability for the Poisson model occurs for larger mean values. This is because in the \textit{easy} scenario the large overdispersion in the training data is compensated by a large Poisson rate such that there are no Poisson samples with low mean and variance. 

These predictions are confirmed by the exact posterior probabilities calculated for the test data set, which are plotted as red circles on top of the MDN predictions in figure \ref{fig:network_visualization}. The matching color code shows that for both scenarios the MDN predictions are in close agreement with the exact posterior probabilities. 
\begin{figure}
\centering
\includegraphics[width=\textwidth]{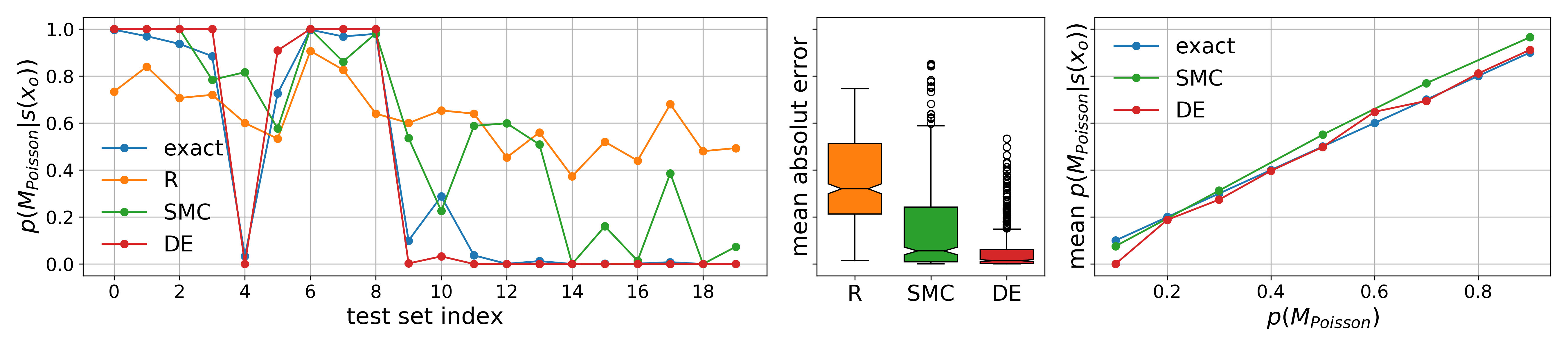}
\caption{Comparison of model comparison methods. The predictions of the basic rejection sampling algorithm (R, orange), the sequential Monte Carlo algorithm (SMC, green) and the posterior density estimation method proposed here (DE, red) are shown for a set of test samples in the left subplot. In the middle subplot the comparison of absolute error between the predictions and the exact posterior probabilities is shown. In the right subplot the prior probability for the Poisson model during training is plotted against the posterior probability for the SMC (green) and the DE method (red).}
\label{fig:prior_checks}
\end{figure}

A more precise comparison of the predicted and the exact posterior probabilities is presented in figure \ref{fig:prior_checks}. I compared the predictions of the basic rejection sampling algorithm (R), the commonly used SMC algorithm and the posterior density estimation (DE) method against the exact posterior probabilities for all test samples. An exemplary plot of the predictions of a subset of test samples (figure \ref{fig:prior_checks}, left) shows that the basic rejection algorithm is usually far away from the exact posterior probability; the SMC algorithm performs well and is sometimes very close to the exact solutions; the DE method is even closer to the exact posterior for most samples, however, it tends to overestimate with predictions close to either 0 or 1. Averaged over the set of test samples the absolute error to the exact posterior probabilities is the lowest for the DE method (figure \ref{fig:prior_checks}, middle).  

Additionally, I tested how the predicted posterior probabilities depend on the prior over model indices. This test is important for the use case, where no ground-truth is available. For the example problem we find that the posterior probability of a given model averaged over a large test set corresponds to the prior probability of the model for both the DE and SMC method (figure \ref{fig:prior_checks}, right). 

\subsection{Approximating the posterior over model parameters}
The MDN for estimating the posterior over models reliably predicts the underlying model posterior probability given observed counting data. I trained another MDN that estimates the posterior over the model parameters, given the predicted model index and the observed data. As before, because I am working with the example problem I have access to the exact posteriors over parameters and can validate the estimated posteriors directly. To this end, I generated $500$ model parameter test samples $\xi_o^i$ from the prior, generated the data $x_o^i$ and calculated the predicted posterior $q_{\phi}^i(\xi | x_o^i)$ as well as the exact posterior $p^i(\xi | x_o^i)$. The posterior of the Poisson model is one-dimensional with $\xi = [\lambda]$, whereas the posterior of the NB model is two-dimensional with $\xi = [k, \theta]$. 
\begin{figure}
\centering
\includegraphics[width=\textwidth]{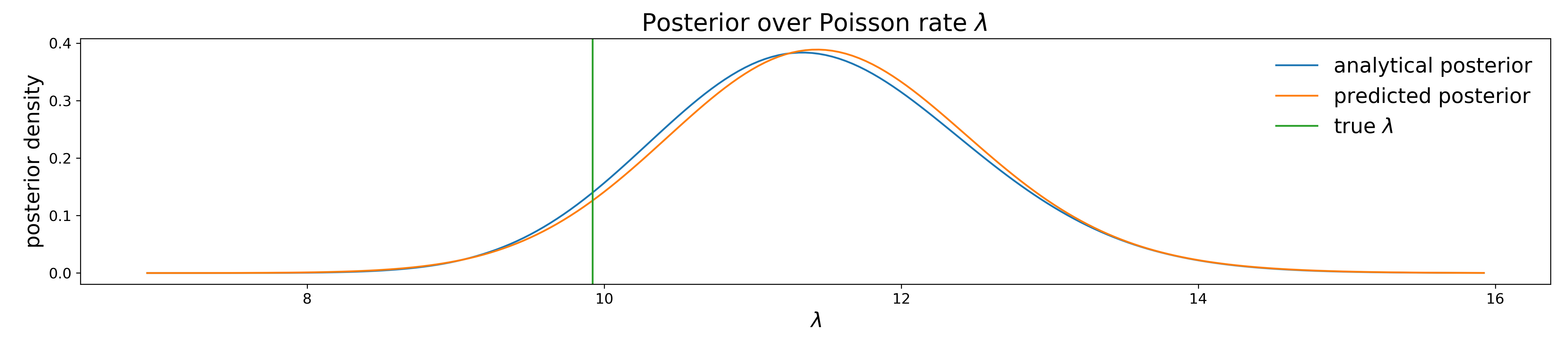}
\caption{The exact and the estimated posterior probability mass function plotted for a range of values for $\lambda$. The true $\lambda$ parameter underlying the data is plotted as a vertical line in green.}
\label{fig:posterior_example}
\end{figure}

\subsubsection{Poisson model}
First, I present the results for the posterior over the Poisson rate $\lambda$. Because the Poisson distribution is in the exponential family and I used the conjugate Gamma prior on the $\lambda$ parameter, the posterior is Gamma distributed. In figure \ref{fig:posterior_example} the resulting posteriors are shown for an arbitrary test sample; for this test sample the exact posterior is estimated very accurately by the MDN. 
\begin{figure}
\centering
\includegraphics[width=\textwidth]{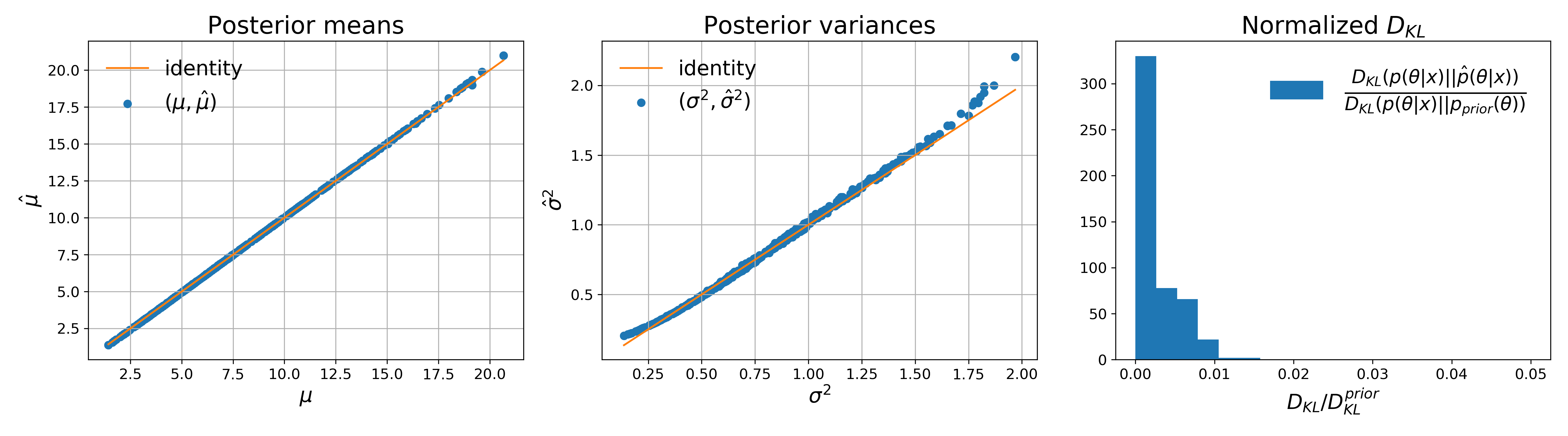}
\caption{Validation of the estimated posterior against the exact posterior. In the left and middle plot the posterior means and variances of the exact and the estimated posterior are plotted against each other for every test sample. The histogram on the right shows the $D_{KL}$ between the posteriors normalized with the $D_{KL}$ between the exact posterior and the prior.}
\label{fig:posterior_checks1}
\end{figure}

For every test sample I calculated the posteriors and inspected the means, the variances and the Kullback-Leibler divergence ($D_{KL}$) in comparison with the exact posterior. The posterior means fit very well for all test samples (figure \ref{fig:posterior_checks1}, left), and the posterior variances show only small deviations when they are relatively small or relatively large (figure \ref{fig:posterior_checks1}, middle). The $D_{KL}$ between the exact posterior and the estimated posterior normalized with the $D_{KL}$ between the exact and the prior is close to zero for most test cases (figure \ref{fig:posterior_checks1}, right). This indicates that the estimated posteriors are much closer to the exact posterior than the prior. 
\begin{figure}
\centering
\includegraphics[width=\textwidth]{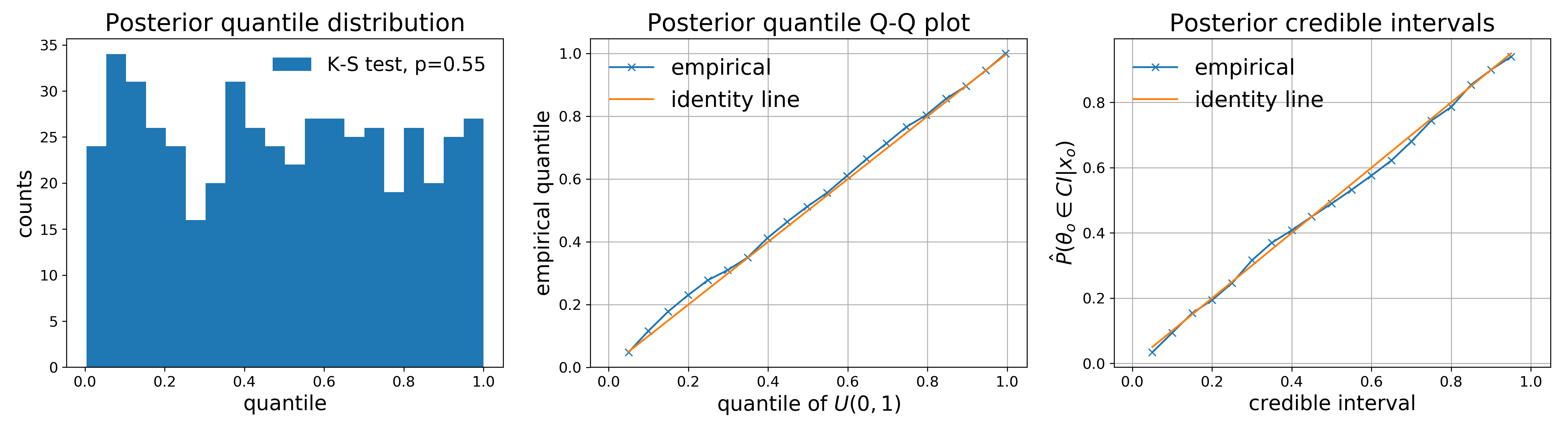}
\caption{Validation of the estimated without the exact posterior: Left, the histogram of quantiles of the test parameters $\lambda_o^i$ under the estimated posterior; middle, the Q-Q plot of the quantile distribution and a uniform distribution; right, the comparison of the estimated probability of a test parameter $\lambda_o^i$ to fall into a credible interval and the corresponding credible mass.}
\label{fig:posterior_checks2}
\end{figure}

The approximated posterior was validated independently of the exact posterior as well. For a method that estimates the posterior accurately the distribution of quantiles of $\lambda_o^i$ under the posterior $\hat{p}^i$ is uniform (for details, see section \ref{sec:posterior_validation}). This seems to be the case here because the Q-Q plot between the observed quantiles and the quantiles of a uniform distribution is very close to the identity line (figure \ref{fig:posterior_checks2}, middle). Finally, I estimated the probability of a $\lambda_o^i$ to fall into a given credible interval of the estimated posterior. For an accurate posterior distribution, this probability corresponds to the mass in the credible interval. And indeed, for this test set, the estimated probabilities and the credible interval masses lie very close to the identity line (figure \ref{fig:posterior_checks2}, right). 

\subsubsection{Negative binomial model}
The validation of the parameter posterior of the NB model is more elaborate because the posterior is two-dimensional. As a first step I examined the one-dimensional marginal posteriors $p(k | x_o)$ and $p(\theta | x_o)$. 
\begin{figure}[H]
\centering
\includegraphics[width=\textwidth]{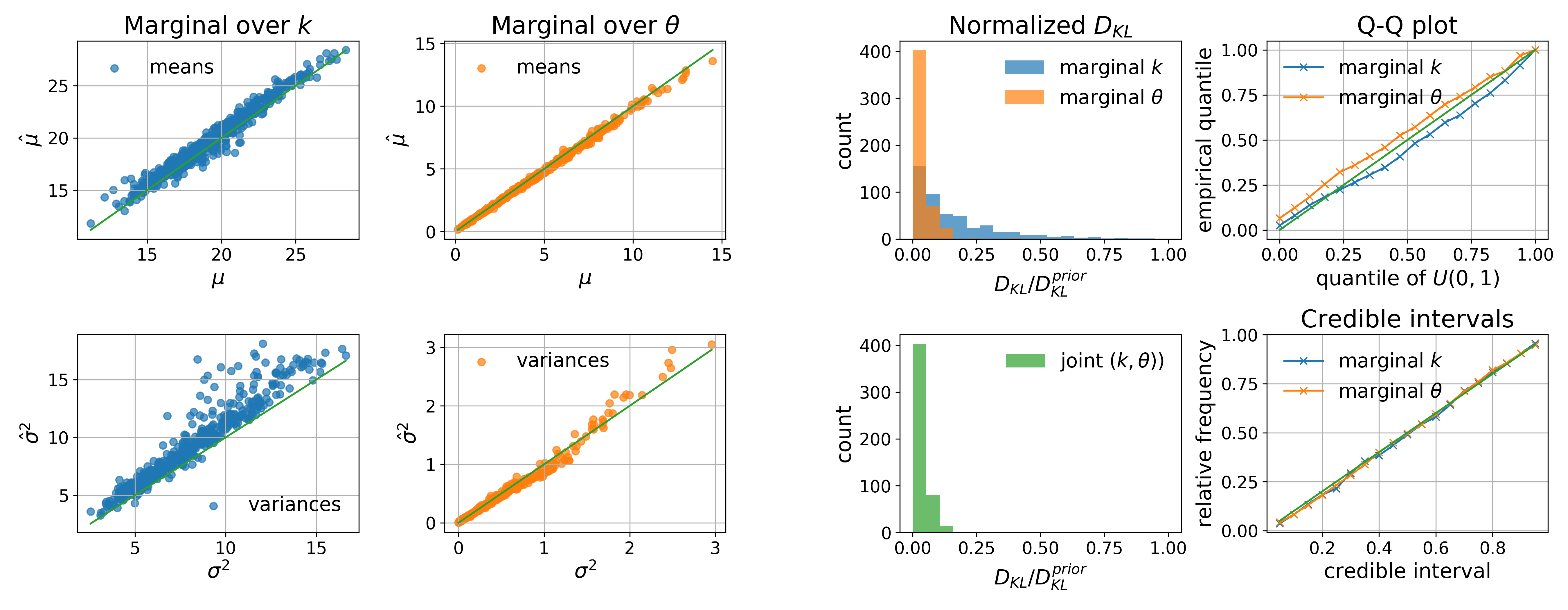}
\caption{Comparison of the marginals of the estimated posterior and the exact posterior. The marginal means and variances are plotted against each other for every test sample (first and second column). The histograms in the third column show the normalized $D_{KL}$ between the marginal posteriors (top) and the joint posteriors (bottom). And in the fourth column are plotted the Q-Q plots and the credible interval probabilities of each marginal.}
\label{fig:nb_posterior_marginals}
\end{figure}

As before, I compared the means and variances, calculated the normalized $D_{KL}$, the posterior quantiles and the credible interval probabilities for the two marginals separately. Again, the means are well fitted, whereas the variances are over-estimated for some test samples (figure \ref{fig:nb_posterior_marginals}, top left and middle). The normalized $D_{KL}$ is close to zero for most $\theta$-marginals; for the  $k$-marginals there are a couple of test samples for which it is close to one, indicating that the marginal is as far away from the exact one as the prior (figure \ref{fig:nb_posterior_marginals}, top right). Overall, however, the marginals seem to be estimated accurately as the Q-Q plot of each marginal indicates a uniform quantile distribution and the credible interval probability is close to the underlying credible mass (figure \ref{fig:nb_posterior_marginals}, right column). One can also observe that the normalized $D_{KL}$ between the joint estimated posterior and the exact posterior is close to zero for most test samples. 

\begin{figure}[H]
\centering
\includegraphics[width=\textwidth]{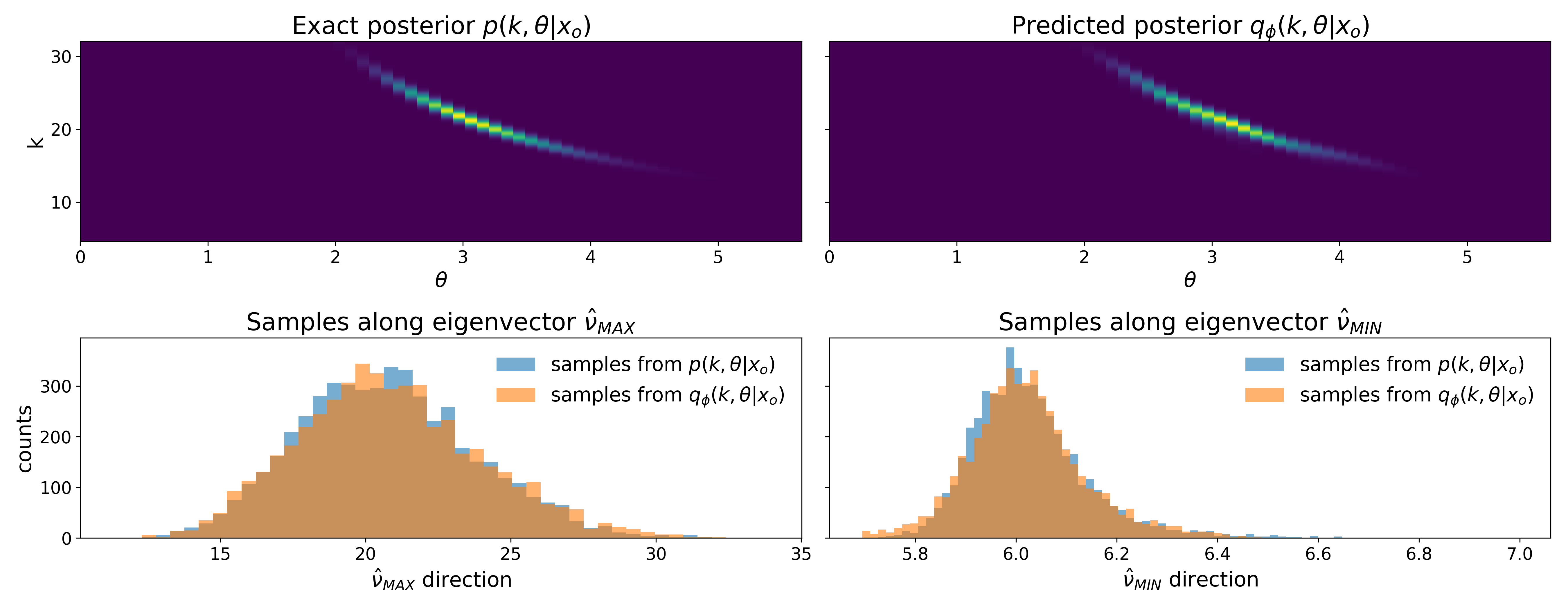}
\caption{Visualization of the posterior of the negative binomial model: top row, the probability density function of the two-dimensional posterior for the exact posterior (left) and the estimated posterior (right), bottom row, histograms of samples from both the exact posterior and the predicted posterior projected along the eigenvectors $\nu_{MAX}$ (left) and $\nu_{MIN}$ (right) of the estimated posterior.}
\label{fig:nb_posterior_example}
\end{figure}

To validate the joint posterior in two dimensions I investigated the eigenvalues $\lambda_i$ and eigenvectors $\nu_i$ of the covariance matrix of the estimated and the exact posterior. One can quantify how well the covariance is estimated by projecting samples from the exact and from the estimated posterior along the estimated eigenvectors of largest and of smallest variance, $\hat{\nu}_{MAX}$ and $\hat{\nu}_{MIN}$. This is illustrated in figure \ref{fig:nb_posterior_example} for an arbitrary test sample $([k, \theta], x_o)$.  

The non-Gaussian shape of the exact posterior seems to be well approximated by the mixture of Gaussians (figure \ref{fig:nb_posterior_example}, top row). Likewise, the distribution of samples from the exact posterior along the marginal directions of $\hat{\nu}_{MAX}$ and $\hat{\nu}_{MIN}$ resembles that one of the samples from the estimated posterior (figure \ref{fig:nb_posterior_example}, bottom row). 
\begin{figure}
\centering
\includegraphics[width=\textwidth]{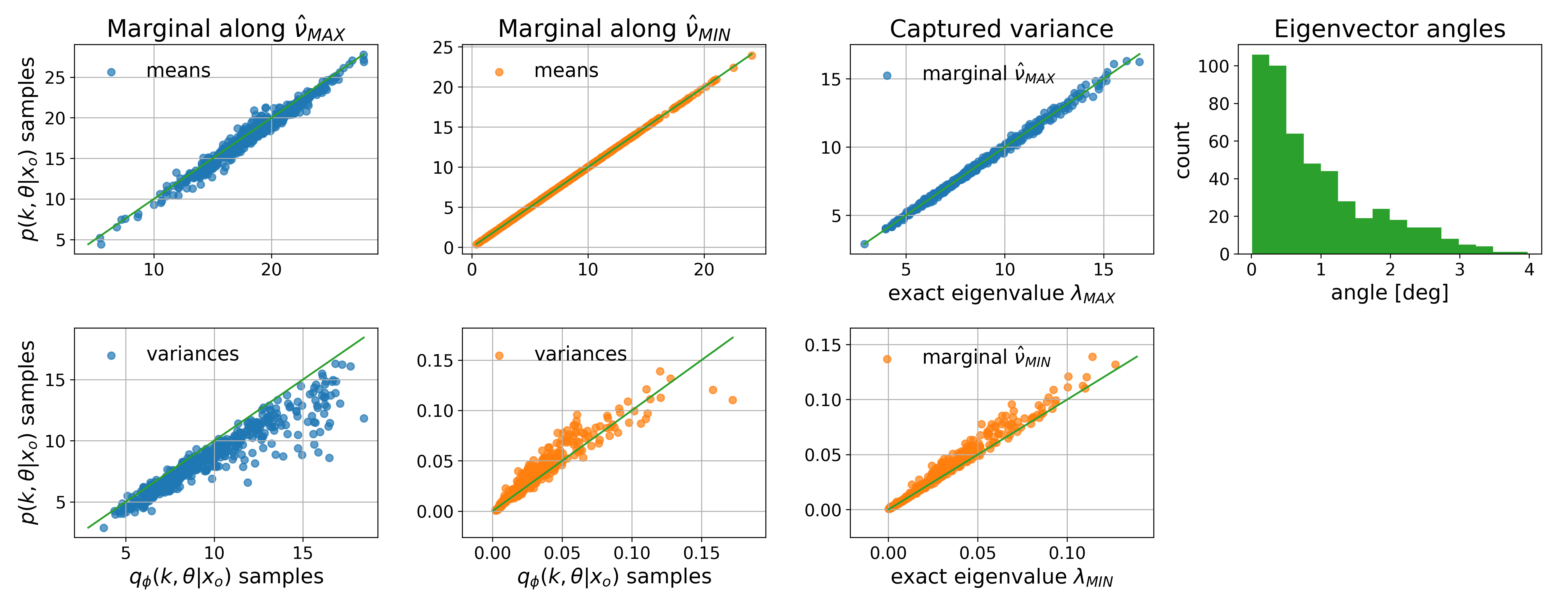}
\caption{Validation of the two-dimensional posterior of the NB model. The first and second column show the sample means (top row) and variances (bottom row) of the samples from the exact and the estimated posterior projected along the eigenvector directions of largest (blue) and smallest (orange) variance. In the third column the projected variance of exact samples along the eigenvector marginals of the estimated posterior is plotted against corresponding largest (blue) and smallest (orange) eigenvalue of the exact posterior. The fourth column shows a histogram of the angles between the exact and the estimated eigenvectors.}
\label{fig:nb_posterior_eigenvectors}
\end{figure}

To test whether this holds throughout the set of test samples I compared the means and variances of the exact samples and the estimated samples projected along the eigenvectors. I find that the means match very well (figure \ref{fig:nb_posterior_eigenvectors} first row, left), whereas the variances are slightly overestimated along $\hat{\nu}_{MAX}$ and underestimated along $\hat{\nu}_{MIN}$ (second row, left). Still, this indicates that the directions of largest and the smallest variance in the estimated posterior also appear in the samples from the exact posterior. 

To further quantify how much variance of the true data is captured by the eigenvectors of the estimated posterior, I plot the variance of the samples from $p(k, \theta | x_o)$ projected along $\hat{\nu}$ against the actual variance those samples have under $p(k, \theta | x_o)$, i.e. against the eigenvalue $\lambda$ (figure \ref{fig:nb_posterior_eigenvectors}, third column). Along $\hat{\nu}_{MAX}$ the projected sample variance and the corresponding eigenvalue $\lambda_{MAX} $ of the exact posterior match closely (third column, top), whereas along $\hat{\nu}_{MIN}$ the projected variance seems to be slightly larger than the actual variance for some test samples (third column, bottom). Nevertheless, the angles between the eigenvectors are very small for most test samples (figure \ref{fig:nb_posterior_eigenvectors}, top right), indicating a very good estimation of the posterior covariance. 

\subsection{Summary}
Overall, the proposed method for estimating the posterior over model indices and the posterior over model parameters appears to work accurately for the example problem. Most importantly, the posterior probabilities of the models are estimated accurately (figure \ref{fig:network_visualization}). Additionally, I find that the posterior over the model parameters of the Poisson model is estimated very accurately (figure \ref{fig:posterior_checks1}) and that the two-dimensional posterior of the NB model is estimated accurately for most test samples (figure \ref{fig:nb_posterior_marginals}).  
\chapter{Model comparison on ion channel models}
\label{ch:ion_channel_example}
Ion channels play a key role in neuronal dynamics \citep{koch2000}. Mechanistic models of ion channels are therefore a central object of computational neuroscience. A large number of families, types and subtypes of ion channels has been identified by experimental research and in parallel, a large number of computational models exists. While these models have been organized in a centralized manner, for example in a publicly available data base of computational models \citep{hines2004}, the difficulty to compare them systematically still remains. This is because no common basis for comparison is available. 

Recently, \cite{podlaski2017} made the effort to systematically compare and categorize a large number of published voltage gated and calcium gated ion channel models and set up an ion channel genealogy (ICG) \citep{ICGenealogy2016}. Their comparison is based on ion channel model meta data from the literature and on ion channel model kinetics in response to standardized voltage clamp protocols. Using the ICG API it is possible to either categorize new ion channels models within the set of existing models, or to find ion channel models that fit to current traces recorded experimentally. 

The method for approximate Bayesian model comparison defined in chapter \ref{ch:general_methods} has the potential to further simplify the process of selecting among competing models and of fitting parameters. I therefore choose the ion channel model comparison as a use-case scenario for the proposed method. From the ICGenealogy I select two published ion channel models to compare them based on the standardized voltage clamp protocol proposed by \cite{podlaski2017}. Like in the example problem illustrated in chapter \ref{ch:toy_example}, I first approximate the posterior over models and then, given a model selected based on the predicted posterior probability, I approximate the posterior over parameters for each model. The main difference to the example problem in chapter \ref{ch:toy_example} is that the models are more complex and that there is no ground-truth posterior available. 

The remainder of this chapter will introduce the channel models in more detail, explain the voltage traces and summary statistics of the ion channel model data and the training and testing procedures. The corresponding results are shown in section \ref{sec:channel_results}. 

\section{Methods}
\subsection{Ion channel models}
The ion channel models compared here are taken from an article by \cite{pospischil2008}. The authors develop a set of simple yet flexible models that are able to fit a diversity of cortical and thalamic single neuron recordings. The models are based on voltage dependent currents modeled with a basic set of ion channel variants: sodium ($I_{Na}$) and potassium ($I_{K_d}$) currents for action potential generation, a slow voltage-dependent potassium current ($I_{K_s}$) for spike-frequency adaptation, a high-threshold calcium current ($I_T$) and a low-threshold calcium current ($I_L$). Together, they govern the dynamics of the membrane voltage described by
\begin{align}
C_m \frac{dV}{dt} &= -g_{leak}(V - E_{leak}) - I_{Na} - I_{K_d} - I_{K_s} - I_T - I_L.
\end{align}
The current equation is given by 
\begin{align}
I_j &= \bar{g}_j \; m^M \; h^N \; (V - E_j), 
\end{align}
that is, the product of peak conductance $\bar{g}_j$, the activation ($m$) and inactivation ($h$) variables and the driving force $(V - E_j)$. 

The ion channel models differ in the dynamics on $m$ and $h$. For the model comparison I will focus on the delayed-rectifier potassium channel $K_d$ and the slow non-inactivating potassium channel $K_{s}$. These two channels are typically associated with distinct activity features, making the distinction between models easy. However, there might be parameter ranges for which such currents have similar responses, making the problem of model comparison harder. In what follows, I will perform model comparison for parameter sets for which the models responses are clearly distinct.

\subsubsection{Delayed rectifier $K^+$ channel}
The dynamics of the delayed-rectifier $K^+$ current are introduced here as a modification of the original Hodgkin-Huxley equations made by \cite{Traub1991}. The channel is voltage dependent, does not contain inactivation, has relatively slow dynamics and mediates re-polarization of the membrane potential through the outflow of $K^+$  after the upstroke of the action potential \citep{hodgkin1952}. Therefore it is named \textit{delayed-rectifier}. The current equations and the dynamics are the following: 
\begin{align}
I_{K_d} &= \bar{g}_{K_d} \; n^{M} \; (V - E_K) \\
\frac{dn}{dt} &= \alpha_n(V) (1 - n) - \beta_n(V) n \\ 
\alpha_n(V) &= \frac{R_{\alpha} (V - V_T - th_{\alpha})}{\exp[-(V - V_T - th_{\alpha})/q_{\alpha}] - 1} \\ 
\beta_n(V) &=R_{\beta} \; \exp[-(V - V_T - th_{\beta}) / q_{\beta}], 
\end{align}
where the peak conductance is set to $\bar{g}_{K_d}= 5 \; \text{mS/cm}^2$ and the reversal potential to $E_K=-90$mV. While these two parameters are biologically constrained, there is a number of free model parameters that can take different values dependent on observed data to be reproduced by the model. In \cite{Traub1991} and \cite{pospischil2008} those parameter were set as follows: $M=4$, $R_{\alpha}=0.032$, $V_T=-63$, $th_{\alpha}=15$, $q_{\alpha}=5$, $R_{\beta}=0.5$, $th_{\beta}=10$ and $q_{\beta}=40$. I define this parameter configuration as the ``ground-truth'' in the following. 
\subsubsection{Slow non-inactivating $K^+$ channel}
The slow non-inactivating potassium current as used in \cite{pospischil2008} was described originally by \cite{Yamada1989} as follows: 
\begin{align}
I_{K_s} &= \bar{g}_{K_s} \; p^{M} \; (V - E_K) \\
\frac{dp}{dt} &= (p_{\infty}(V) - p) / \tau_p(V) \\ 
p_{\infty}(V) &= \frac{1}{1 + \exp[-(V + th_p)/q_p]} \\ 
\tau_p(V) &= \frac{\tau_{max}}{R_{\tau}\exp[(V - th_p)/q_{\tau}] + \exp[-(V - th_p)/q_{\tau}]} , 
\end{align}
where $\bar{g}_{K_s}= 0.004 \; \text{mS/cm}^2$ and activation time constant $\tau_{max}=4$s. Again, the latter two parameters are biologically well constrained, whereas one can fit the five remaining parameters to observed data. The values chosen by \cite{pospischil2008} and \cite{Yamada1989} are the ``ground-truth'' for the posterior fitting demonstrated later: $M=1$, $th_p=35$, $q_p=10$, $R_{\tau}=3.3$ and $q_{\tau}=20$. 

\subsection{Voltage-clamp protocols} One goal of the ICGenealogy is to compare channel models based on their dynamic responses to voltage-clamp protocols. The protocols were designed to investigate the whole range of dynamics of the model \citep{hodgkin1952, Willms1999} without taking into account the model equations explicitly. \cite{podlaski2017} used an individual set of five protocols for every ion type, testing the gating characteristics of the ion channel: 
\begin{itemize}
\item 
a single voltage step to capture the \textbf{activation} kinetics
\item 
a varying voltage step followed by a fixed voltage step to capture the \textbf{inactivation}
\item 
a single high voltage step followed by varying voltage steps for the \textbf{deactivation}
\item 
the membrane voltage during a series of regular action potentials
\item 
four up and down ramping voltages with different slopes.
\end{itemize}

I will take this approach as the basis for the channel model comparison: I use the voltage-clamp protocols defined for the $K^+$ current and simulate the response to the five different protocols through the channel equations defined above. The responses are normalized with the corresponding maximal conductance to be comparable across models and subsampled to 512 samples. Because the protocols consist of voltage traces of $c=12$ different step amplitudes, this results in $c$ current traces that are arranged in a single vector, one per protocol. 

\subsection{Summary statistics} To summarize this data for the posterior estimation routines and to remove the time dependence, I perform principal component analysis (PCA) across the time dimension as proposed by \cite{podlaski2017}. In particular, the goal of the PCA is to relate the current traces simulated through the model to the dynamics in the entire family of $K^+$ ion channel models. Therefore, the PCA is run beforehand on the set of the current responses of all $N=1243$ $K^+$ ion channel models currently available through the ICG API \citep{ICGenealogy2016}. This is done for every voltage trace protocol separately. Then, the data matrix for the PCA consists of the channel models in rows ($N$) and the appended current traces in the columns ($c \cdot 512$). The first five principle components (PC) are taken as basis functions for the dimensionality reduction, retaining about $99\%$ of the total variance. 

For a given set of simulated current traces the summary statistics are calculated using a least squares approach. The five PCs that were calculated beforehand are used as the basis functions $X$ for fitting the current traces $\mathbf{y}$ as follows
\begin{align}
X \beta &= \mathbf{y}, \\ 
\hat{\beta} &= (X^T \; X)^{-1} \; X^T \mathbf{y}
\end{align}
and the estimated coefficients $\hat{\beta}$ are taken as summary statistics. Given five basis functions and five voltage protocols this results in 25 summary statistics per data sample. 

\subsection{Training procedure}
The mixture-density networks (MDN) for learning the mapping from observed data to posteriors over model indices and over model parameters are trained with a data set of model indices, model parameters and summary statistics: $\{m_n, \theta_n, s(x_n)\}_{n=1}^N$. 

To generate the training data set, equal prior probability is assigned to each model $\mathcal{M}_i$, and uniform priors are chosen for the model parameters $\theta^i$ in a range around the ground-truth ($GT$) value defined above: 
\begin{align}
p(\mathcal{M}_{K_d}) &= p(\mathcal{M}_{K_s}) = 0.5 \\ 
p(\theta^i) &= \mathcal{U}(0.3 * \theta^i_{GT}, \;  1.3 * \theta^i_{GT}). 
\end{align}
A training data set of $N=100000$ samples is generated. The overall training data set of model indices and summary statistics, $(m_n, s(x_n))$, is used to train a MDN $q_{\psi}(m | x)$ to approximate the model index posterior. The MDN consists of a single hidden layer with 10 units and is trained with a batch size of $N/100$ for 10 epochs. 

For the model parameter posteriors I train separate MDNs for each channel model using the corresponding training data $(\theta_n, s(x_n))$. The one for the delayed-rectifier model $K_d$, $q^{K_d}_{\phi}(\theta | x)$, is an $8$ dimensional mixture of Gaussians (MoG). The MoG for the $K_s$ model is 5 dimensional. The training parameters are the same for both: 2 hidden layers holding 30 units, 3 mixture components in the mixture of Gaussians, a batch size of $N/100$ and 100 training epochs.  

Throughout training, a learning rate of $\eta=0.01$ and Adam optimization with default parameters is used. 

\subsection{Testing procedure}
The posterior approximation is evaluated with a testing data set generated in the same way but independently from the training data set. Because there is no ground-truth posterior available for this use-case example, the posterior validation methods are limited. 

For the discrete posterior over model indices, the consistency of the predictions with changing model prior probabilities is tested. To this end, several testing data sets are generated with different priors on the model classes and the average predicted posterior probability is compared to the prior probability of each model. 

As no ground-truth posterior probabilities are available, the predictions for the posterior over models cannot be validated directly. However, it is possible to compare the predicted posterior probability to the actual underlying model, e.g., by comparing the predicted probability for the $K_d$ model to the underlying model $m_i \in \{ 0, 1\}$ with a cross entropy loss. It is important to note that this is not validating the accuracy with respect to the actual posterior probability given the observed data. However, it allows to compare the performance, in terms of selecting the correct model, to other methods. Therefore, I calculate the cross entropy loss on the test data set and compare it to the loss of the standard rejection sampling procedure and the sequential Monte Carlo (SMC) rejection sampling method on the same test set. 

In case of the continuous model parameter posterior, a first evaluation is the visual comparison of the current traces simulated with the ground-truth parameters and the current traces simulated with the mode or random samples from the estimated posterior. For a more systematic validation, I perform the posterior validation tests as described in the methods section \ref{sec:posterior_validation}: I examine the distribution of posterior quantiles across parameters of the testing data set and I check the relative frequency with which test parameters fall into posterior credible intervals of different widths. 

\section{Results}
\label{sec:channel_results}
Mixture-density networks (MDN) were trained to approximate the posteriors over ion channel models $K_d$ and $K_s$ and over the corresponding model parameters. For MDN training I used a large data set of summary statistics on current traces of the models in response to stereotypical voltages traces. In this section the results are presented. First, the voltage protocols and current traces for both models, then the validation of the posterior over models and finally the results for the posteriors over model parameters of both models. 

\subsection{Voltage protocols and current traces}
\begin{figure}[H]
\centering
\includegraphics[width=\textwidth]{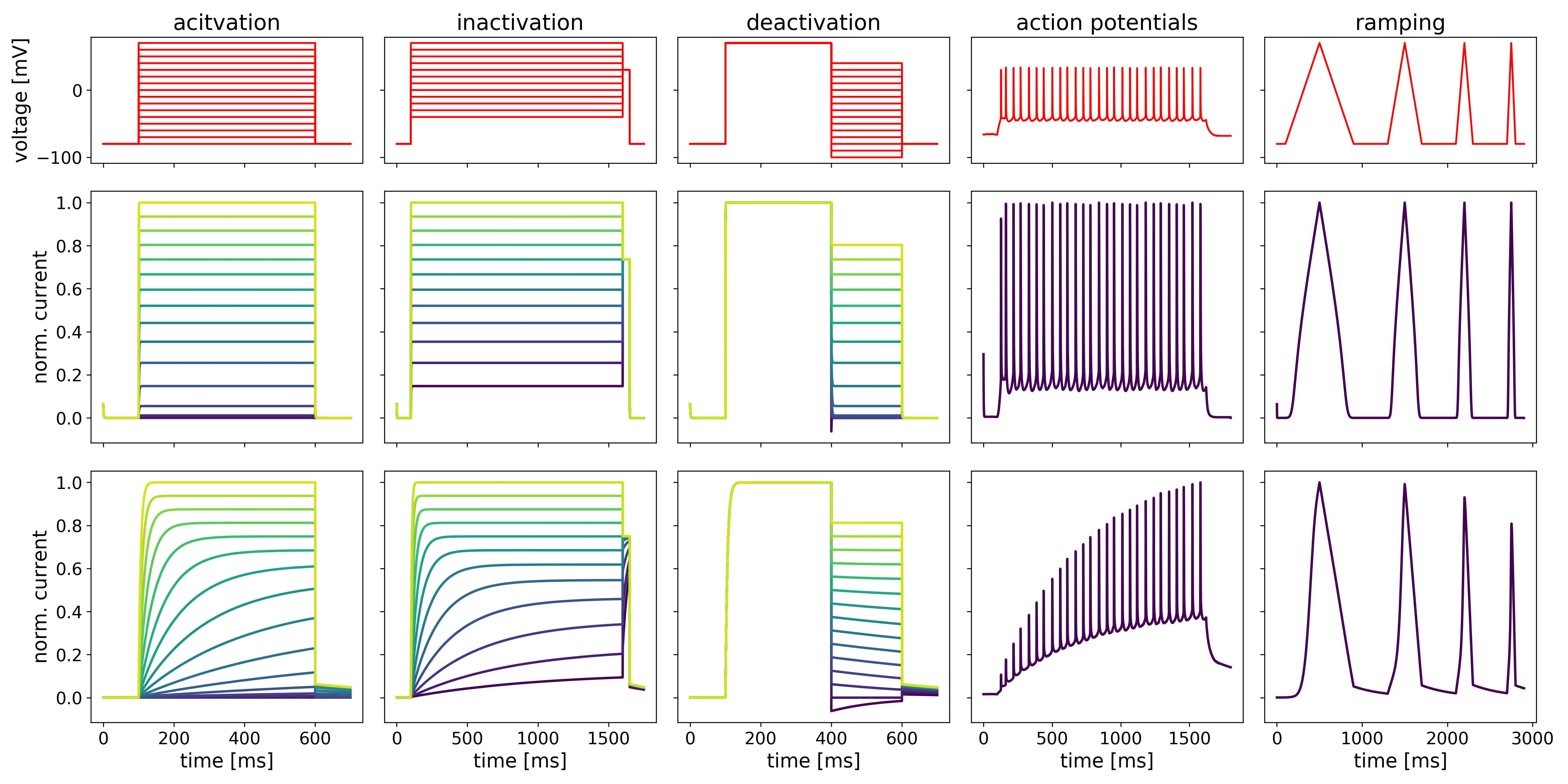}
\caption{Voltage protocols and current responses. The protocols are shown in the subplots of the first row. In the second and third row the corresponding current trace responses of the $K_d$ and $K_s$ ion channel model are shown, respectively. For better comparison the current responses are normalized with their largest value.}
\label{fig:channel_traces}
\end{figure}

The voltage protocols are used to characterize the behavior of the channel models in different dynamics regimes. The comparison of channel model is then based on the current responses to the voltage traces. For the parameters provided in the original article by \cite{pospischil2008} a clear difference between the responses of the $K_d$ and the $K_s$ is visible (figure \ref{fig:channel_traces}). 

For example, one can clearly see the characteristic slow time scale of the $K_s$ channel model for the \textit{activation} and the \textit{inactivation} protocol (figure \ref{fig:channel_traces}, third row left). Additionally, the spike-frequency adaptation effect becomes visible in the \textit{action potential} protocol when the $K_s$ current slowly increases in amplitude. 
\subsection{Approximating the posterior over models}
In the ion channel model comparison example there are no ground-truth posteriors available, therefore the validation of the estimated posterior probabilities is limited. Two tests were performed with a test set of 1000 samples $(m_n, s(x_n))$: the performance compared to standard rejection sampling approaches and the dependence on the prior over models. 

The comparison between methods was based on the cross entropy loss between the predicted posterior probabilities and the underlying model indices. The loss was calculated for the posterior density estimation (DE) method, the basic rejection sampling method (R) and the sequential Monte Carlo (SMC) method. 
\begin{figure}[H]
\centering
\includegraphics[width=\textwidth]{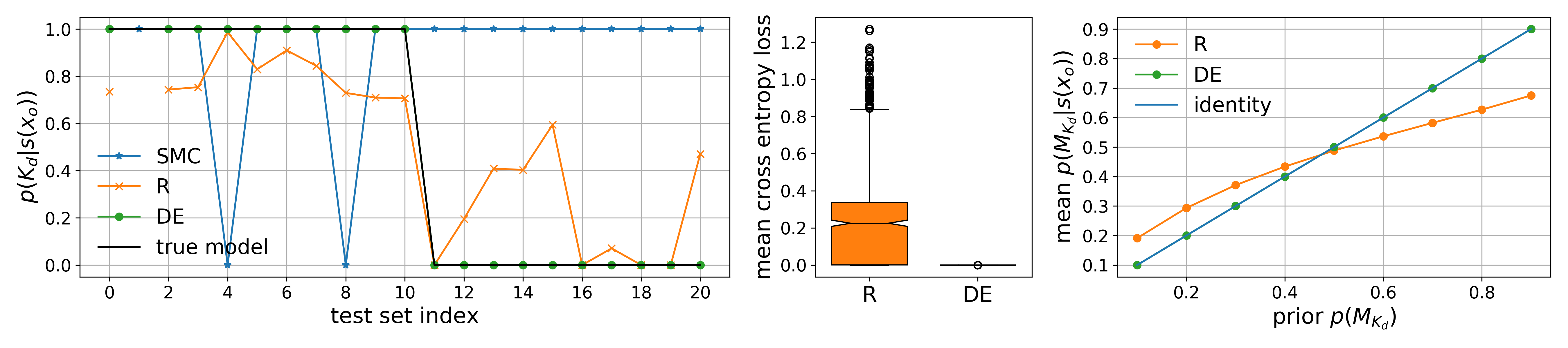}
\caption{Validation of the posterior over channel models: Left, the predicted posterior probability for a subset of test samples for the rejection sampling (R), SMC and the posterior density estimation method (DE) (note that the true model $\in \{0, 1\}$ is plotted on the same scale as the posterior probability); middle, the mean cross entropy loss over the entire test set; right, the prior probability of model $K_d$ in the test set plotted against the average predicted posterior probability.}
\label{fig:channel_priorcheck}
\end{figure}

The visualization of a subset of predictions from the test set (figure \ref{fig:channel_priorcheck}, left) indicates that for this test set and selection of algorithm hyper-parameters the rejection sampling approach performed reasonably with some exceptions, whereas the SMC method was not able to recover the underlying model. The posterior density estimation method performs very well. Accordingly, the mean cross entropy loss on the entire test is smallest for the this method (figure \ref{fig:channel_priorcheck}, middle, SMC is not shown). 

The second test investigated the dependence of the predicted posterior probabilities on the prior distribution of models in the test set. To this end, I calculated the average posterior probability over a test set for different prior probabilities on the two models. The results show that for the posterior density estimation approach (DE) the posterior reflects the prior probability of each model well, e.g., if the prior for the $K_d$ model was low, then the posterior probability for this model averaged over the test data set was low as well (figure \ref{fig:channel_priorcheck}, right). The rejection sampling approach (R) becomes less accurate when the prior probability are close to 0 or 1 for one of the two models. 

\subsection{Approximating the posterior over model parameters}
Two separate MDNs were trained to approximate the posterior over the parameters of each model and the results are presented here separately. First, I predicted the posterior given the ``ground-truth'' parameters from the original paper \citep{pospischil2008}. The visualization of this posterior gives a first impression of the results. Second, the results of forward simulations of the channel models with parameter sampled from the posterior are presented. Finally, for a more systematic validation, the results of the posterior checks using quantiles and credible intervals are shown. 
\subsubsection{Delayed-rectifier $K^+$ channel}
\begin{figure}[H]
\centering
\includegraphics[width=\textwidth]{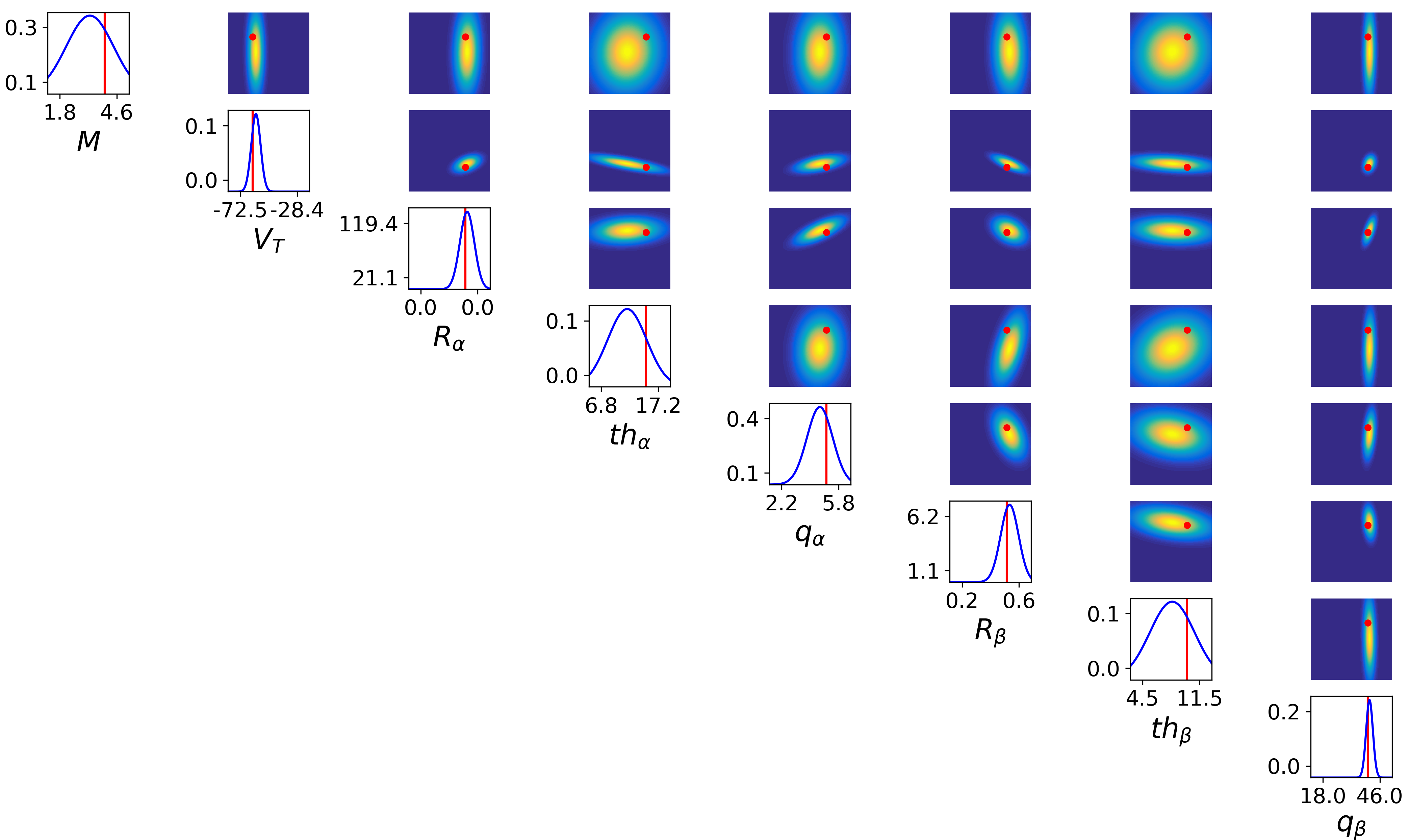}
\caption{Visualization of the eight-dimensional parameter posterior for the $K_d$ model. The marginal \textit{pdf} of each model parameter is plotted on the diagonal and the pairwise marginals across all parameter pairs in the upper triangle subplots. The underlying ground-truth parameters used to generated the observed data are shown as vertical lines in red. The color code represents the posterior density values. Units and detailed ticks are omitted for better visualization: $V_T$ and $th_{\{\alpha, \beta\}}$ are in $mV$, the remaining parameters are unitless.}
\label{fig:channel_posterior_kd}
\end{figure}

The delayed-rectifier $K^+$ channel model has eight free parameters. The posterior was approximated with an eight-dimensional mixture of Gaussians. Its visualization in figure \ref{fig:channel_posterior_kd} shows the one-dimensional marginal posteriors of each parameter on the diagonal and all pairs of two-dimensional marginal posteriors in the upper triangular subplots. The marginal posteriors for $V_T$, $R_{\alpha}$, $q_{\alpha}$, $R_{\beta}$ and $q_{\beta}$ show relatively low variance (relative to the prior variance) and are centered around the ground-truth parameter values. This indicates that those parameters are well constrained by the data and that they are recovered by the posterior approximation. The broader posterior marginals for the remaining parameters $M$, $th_{\alpha}$ and $th_{\beta}$ indicate that those parameter are less constrained by the observed data, e.g., that one can observe the same model output in a range of different values of the parameter $M$, $th_{\alpha}$ and $th_{\beta}$. 

Investigating the shape of the covariance of the two-dimensional marginal posteriors in figure \ref{fig:channel_posterior_kd} gives additional room for interpretation. For example, the negative posterior covariance between $V_T$ and $th_{\alpha}$ indicates that the model output is preserved by simultaneously decreasing $V_T$  and increasing $th_{\alpha}$. Looking at the equations of the $K_d$ model this indeed makes sense because $V_T$ and $th_{\alpha}$ counteract each other. 
\begin{figure}[H]
\centering
\includegraphics[width=\textwidth]{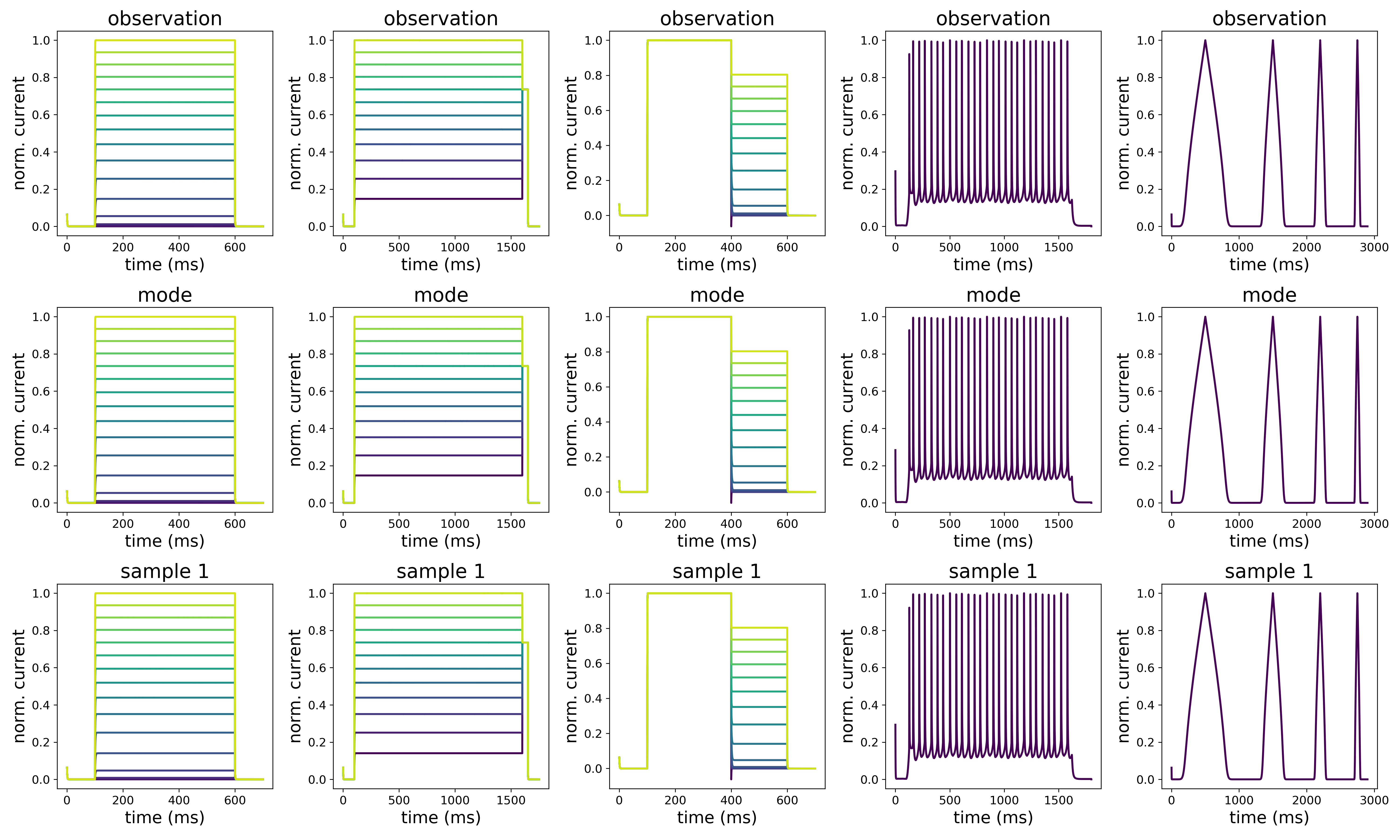}
\caption{Posterior predictive checks for the $K_d$ model posterior. The observed current traces are shown in the first row; the current traces resulting from simulating the posterior mode parameters and a random sample from the posterior in the second and third row, respectively.}
\label{fig:channel_posteriorsamples_kd}
\end{figure}

An additional test for goodness of fit is to compare the observed data with data generated given parameter samples from the learned posterior. The results in figure \ref{fig:channel_posteriorsamples_kd} show that the observed data and the data generated with the parameters of the posterior mode and a random sample from the posterior are very similar.  
\begin{figure}
\centering
\includegraphics[width=\textwidth]{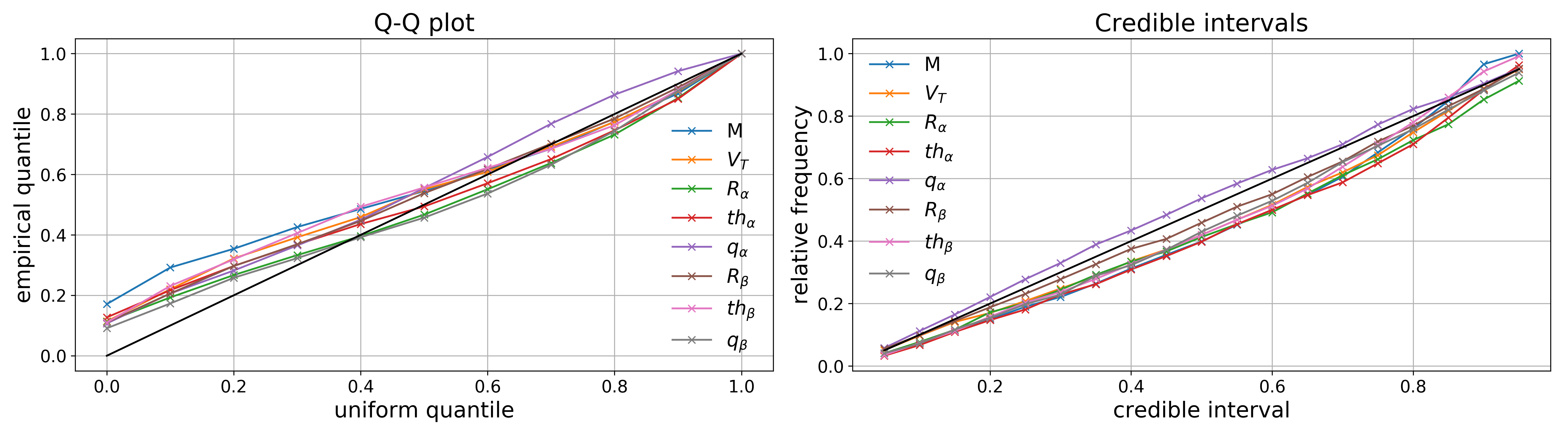}
\caption{Posterior validation with quantiles and credible intervals: The Q-Q plot of the quantiles of the test parameters under the corresponding predicted posterior, and those of a uniform distribution are shown in the left subplot. On the right, the size of different credible intervals ($x$-axis) is plotted against the relative frequency with which parameters from the test set fall into these credible intervals ($y$-axis).}
\label{fig:channel_posteriorchecks_kd}
\end{figure}
 
Finally, to test the performance of the parameter posterior approximation over a large set of test data I investigated the posterior quantiles and credible intervals. Due to the high dimensionality the quantiles and credible interval checks were calculated for every posterior marginal separately. The shapes of quantile distributions are visualized in a Q-Q plot comparing the quantiles of the quantile distribution against the quantiles of a uniform distribution (figure \ref{fig:channel_posteriorchecks_kd}, left). Most marginal quantile distribution are close to the ideal identity line, however, there seems to be a small bias towards smaller quantiles for all of them. This indicates that the posterior marginals are shifted to higher values for some of the test samples. 

The validity of credible intervals was checked for a range of intervals. For visualization, the width of the different credible intervals is compared to the relative frequency with which the underlying parameter falls into the interval. The better the algorithm does at approximating the posterior distribution, the closer the relative frequency is to the identity line. The results in figure \ref{fig:channel_posteriorchecks_kd} (right) show that most marginals are close to the identity line, however, there is a tendency for slightly lower relative frequencies for intermediate intervals. This indicates that for those marginals the estimated posteriors are slightly too narrow, i.e., too confident about the underlying parameter. 
\subsubsection{Slow non-inactivating $K^+$ channel}
\begin{figure}
\centering
\includegraphics[width=0.8\textwidth]{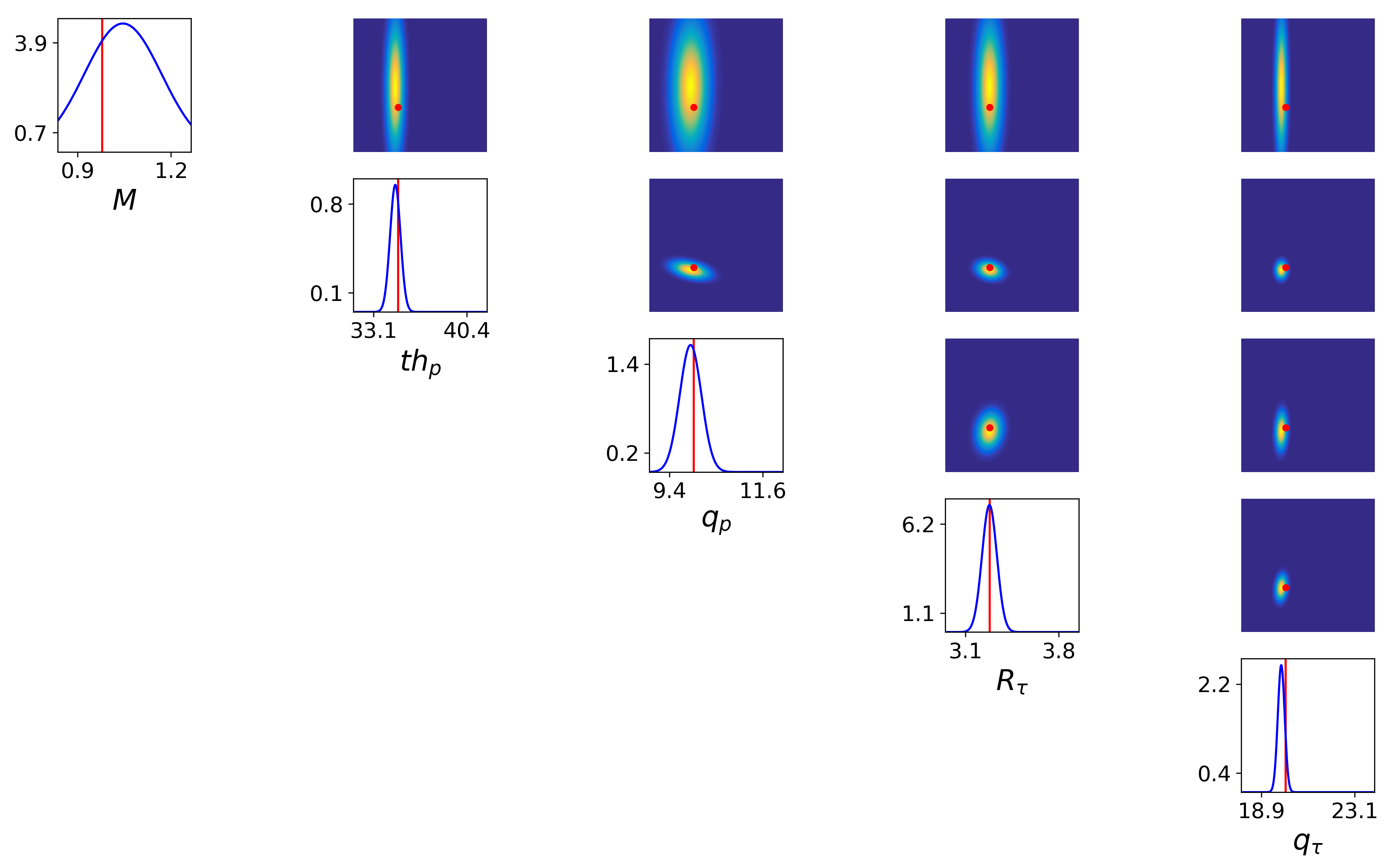}
\caption{Visualization of the five dimensional parameter posterior of the $K_s$ model. On the diagonal the marginal \textit{pdf} of each parameter and on the upper off diagonal subplots the pairwise marginal \textit{pdf}s. The vertical red lines mark the ground-truth parameter values.}
\label{fig:channel_posterior_kslow}
\end{figure}

The parameter posterior for the slow non-inactivating $K^+$ channel model was tested and visualized in the same way. The model has five free parameters so that the approximating posterior is a five-dimensional mixture of Gaussians. The visualization in figure \ref{fig:channel_posterior_kslow} draws a clear picture of how the parameters are constrained by the observed data: the marginals for the parameters $th_p$, $q_p$, $R_{\tau}$ and $q_{\tau}$ are centered on the ground-truth values and have, in relation to the prior distribution very small variance, whereas that of the parameter $M$ has relatively large variance. This indicates that the model is relatively invariant to changes in the parameter $M$. 

\begin{figure}[H]
\centering
\includegraphics[width=\textwidth]{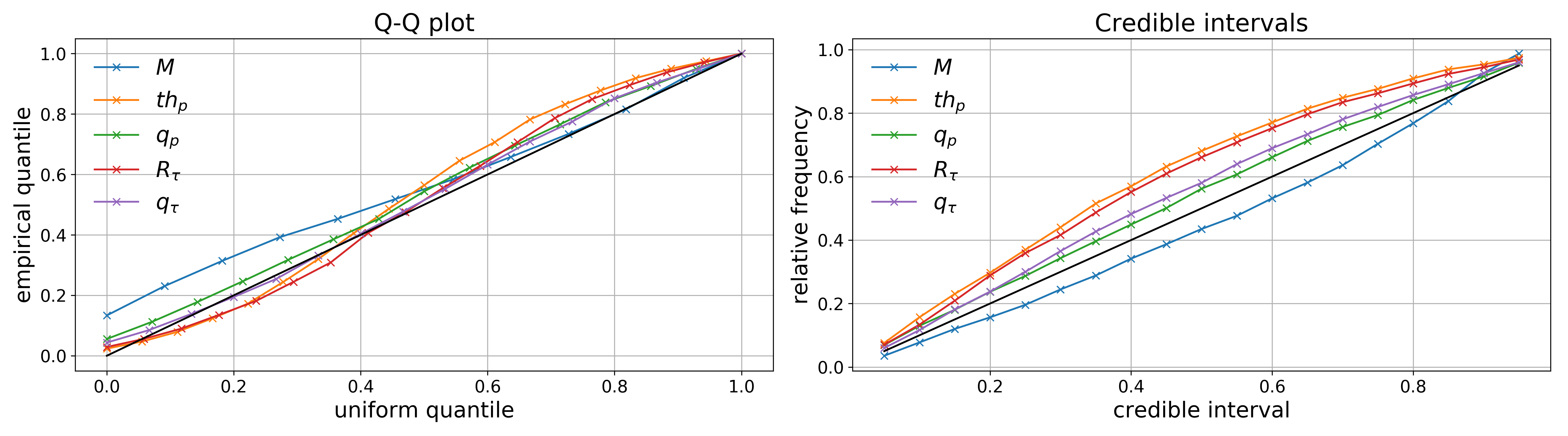}
\caption{Posterior validation with quantiles and credible intervals. The Q-Q plot of the quantiles of the test parameters under the corresponding predicted posterior, and those of a uniform distribution are shown on the left. On the right the size of different credible intervals ($x$-axis) is plotted against the relative frequency with which parameters from the test set fall into these credible intervals ($y$-axis).}
\label{fig:channel_posteriorchecks_kslow}
\end{figure}

The current traces resulting from simulations with the posterior mode and a random sample are very similar to the originally observed data (figure \ref{fig:channel_posteriorchecks_kslow}). This was to be expected, given that the posterior marginals are centered accurately on the ground-truth parameters with low variance relative to the prior. 
\begin{figure}[H]
\centering
\includegraphics[width=\textwidth]{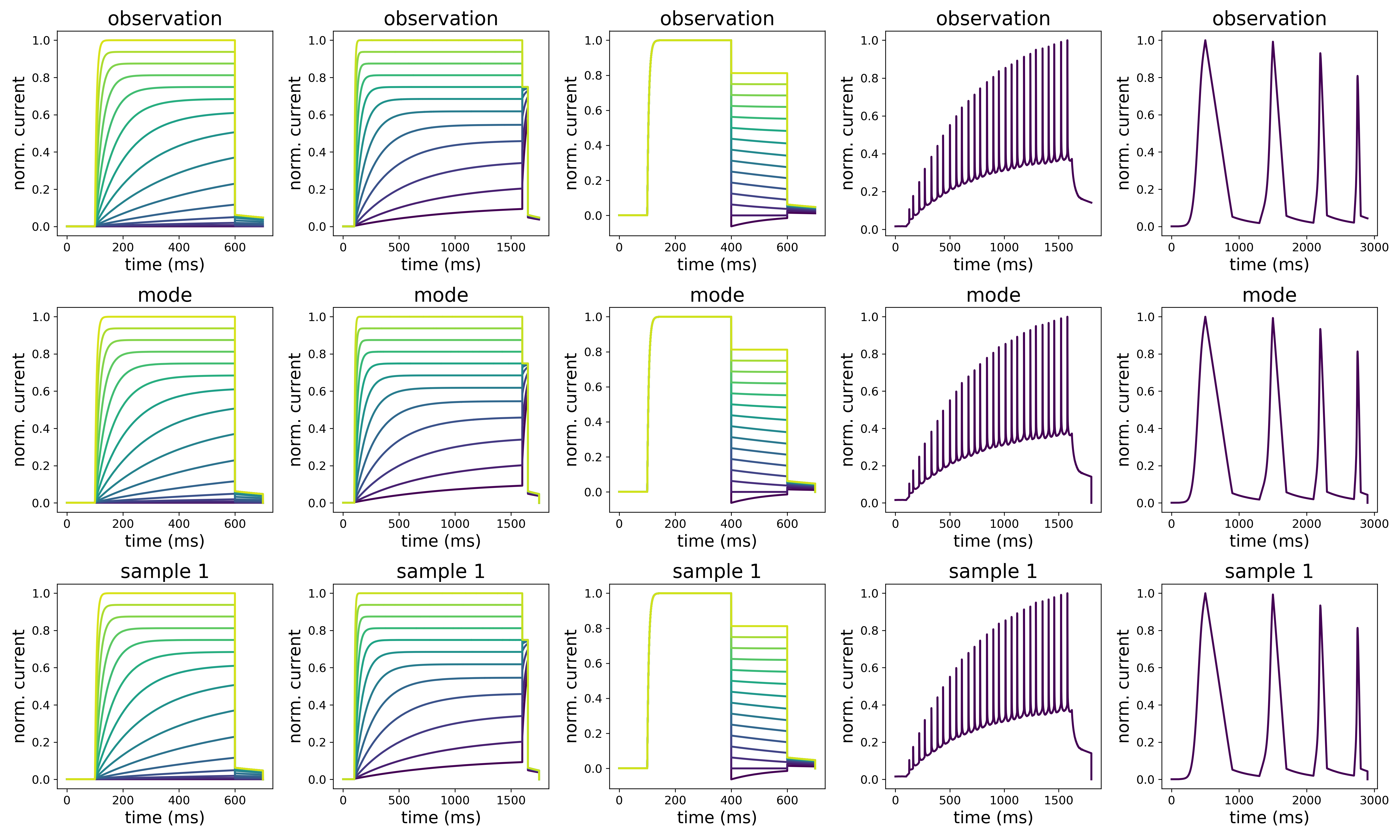}
\caption{Posterior predictive checks for the $K_s$ model posterior. The observed current traces are shown in the first row; the current traces resulting from simulating the posterior mode parameters and a random sample from the posterior in the second and third row, respectively.}
\label{fig:channel_posteriorsamples_kslow}
\end{figure}

The result of the systematic posterior checks over a larger test data are therefore unexpected. The Q-Q plot of the posterior marginal quantiles and a uniform distribution for comparison indicate that there is deviation from the ideal identity line (figure \ref{fig:channel_posteriorchecks_kslow}, left), for example for the marginal of $R_{\tau}$ and $q_{\tau}$. Correspondingly, these two marginals also show larger deviations from the ideal line in the check of the consistency of the credible intervals (figure \ref{fig:channel_posteriorchecks_kslow}, right). The larger values of the relative frequency with which the underlying parameter falls into the credible interval indicates that the posterior marginals are too broad, i.e., the marginal posterior variance is overestimated for these parameters. 

Overall, this shows that although for single examples the ground-truth posteriors are recovered in the sense that the posterior marginals are centered around them, the posterior validation on a larger test set revealed a deficiency indicating that some posterior marginals are not learned consistently.  
\chapter{Discussion}
\lhead{Chapter \thechapter}
The comparison of statistical models in the light of observed data is a central problem in computational sciences. Since classical null-hypothesis significance testing fell into disrepute in the scientific community, Bayesian model comparison has been suggested as an alternative. One limitation of the practical use of Bayesian model comparison is the difficulty of applying it to statistical models used in practice because the likelihoods of the models are often intractable. I presented a new method for performing Bayesian model comparison in cases where the likelihood of the models is not available. In contrast to previous rejection sampling based approaches that quickly become impractical as the number of parameters of a model increases \citep{mckinley2018}, this method is based on posterior density estimation and approximates the posterior in parametric form using all available samples. 

The posterior density estimation approach to model comparison extends recent advances in parameter estimation in approximate Bayesian computation (ABC) \citep{papamakarios2016, lueckmann2017}. These approaches use a mixture-density network (MDN) to estimate the posterior over parameters in closed form by regressing surrogate data that was generated from the underlying model onto the corresponding parameters. They overcome several limitations of the standard rejection sampling algorithms, e.g. they require less samples to accurately estimate the posterior and they perform well on neuroscience models with many parameters and complex parameter priors, as demonstrated in \cite{lueckmann2017}. To carry these improvements to the domain of model comparison I extended the posterior density estimation approach to the estimation of the posterior over model parameters. 

I presented two applications to validate the performance of the proposed method. First, the method was applied to the comparison of a Poisson model and a negative-binomial model. This is a known comparison problem in computational neuroscience for which the ground-truth posteriors are known. The results show that the MDN correctly represents the overdispersion in the data as the crucial difference between the models (figure \ref{fig:network_visualization}). Accordingly, the predicted posterior probabilities were close to the exact posterior probabilities for most test set data points. The overall performance was slightly better compared to a standard SMC algorithm whose predictions were based on the same number of training simulations. 

An important difference to the SMC algorithm was the number of samples required for prediction of the posterior from observed data. While the SMC algorithm is tailored to every observed data point so that it needs to sample from scratch every time a new data point is observed, the proposed method learns the mapping from the data domain to posteriors just once and thereafter predicts a posterior instantly for any new observation. As such it performs \textit{amortized inference}.  If there is no limitation on the number of samples in the SMC algorithm such that it can adapt the rejection threshold to very small values and towards a particular observed data point, then the predicted posterior probabilities eventually become more accurate than those of the posterior density estimation approach. However, this requires that simulations are computationally cheap, which is not the case in practice. In addition, this eliminates the advantage amortized inference. 

In the second example the method was tested in a use-case scenario from computational neuroscience: the comparison of ion channel models. This mimics the situation many scientists encounter: the ion channel models are defined as simulator models through differential equations and it is not clear how to compare them systematically in the light of observed channel data, e.g. current trace responses to voltage protocols. I used summary statistics proposed by \cite{podlaski2017} to learn the mapping from data to the posterior over channel models. The predictions on a test set of simulated data were accurate in the sense that the correct underlying model was predicted with high probability. The comparison to rejection sampling approaches showed that the SMC algorithm was not able to recover the underlying model, whereas the basic rejection sampling algorithm achieved mediocre results. However, the latter required a lot more samples than needed for the MDN training and it was strongly dependent on the selected rejection criterion. Admittedly, the SMC algorithm was applied using an open source toolbox \cite{klinger2017pyabc} with default settings. Thus, it is conceivable to achieve a better result by tuning the algorithm more carefully. Nevertheless, it seems that the rejection sampling approaches are outperformed by the density estimation on a problem with many summary statistics (here 25) and computationally expensive simulations. 

The derivation of the optimization objective for approximating the posterior showed that it is possible to learn both, the posterior over models and the posterior over the parameters for every model, separately. Therefore, in addition to predicting the posterior over models, the posterior over parameters was predicted for both example applications. 

To validate the estimated posteriors, credible intervals and posterior quantile distributions were calculated. In case of the idealized comparison between the Poisson model and the negative-binomial model these demonstrated an accurate approximation. For the case of the posteriors over the parameters of the ion channel models, the visual impression of the posterior was good on individual test cases and predictive checking with samples from the posterior appeared accurate as well. However, the quantile distribution and credible intervals of the marginal posteriors over individual parameters showed deficits for both ion channel models. This indicates that the posterior estimation is not working properly. Reasons for that can be, among others, too little training data for learning the posterior density estimation so that the parameter spaces are not sampled densely enough, or the insufficiency of the summary statistics. 

It was out of the scope of this thesis to investigate this further. However, these issues were addressed in previous work that focuses on the estimation of the posterior over parameters. Therefore, a possible extension of my approach is to apply a more specialized algorithm to the estimation of the posterior over parameters of a selected model, e.g, via density estimation and over multiple rounds with adaptive proposal priors to increase sampling efficiency, like in \citet{lueckmann2017} or via non-linear regression models \citep{blum2010}. 

The problems in estimating the posterior over parameters for the ion channel models indicate an insufficiency in the summary statistics $s(x)$. This can be problematic for the estimation of the posterior probability of the model given the raw data, $p(m | x)$, as pointed out by \cite{robert2011} and explained in section \ref{sec:summary_stats}. Namely, the posterior probabilities estimated based on the summary statistics, $p(m | s(x))$, may in fact be quite different from the actual posterior $p(m | x)$. However, the results presented here showed that throughout the test data set, a posterior probability close to 1 was assigned to the correct ion channel model. Therefore, the information contained in the summary statistics appears to be sufficient to tell the models apart. Yet, in a case where models are more similar so that posterior probabilities are less extreme, the method might produce inconsistent results. To overcome this potential problem, one extension of my approach could be to learn better summary statistics from raw data time series using dimensionality reduction methods as proposed in \cite{blum2013} or by defining the first layers of the MDN as recurrent neural networks applied to the raw data, as proposed by \cite{lueckmann2017} and \cite{jiang2015learning}. 

Overall, the accurate results on the theoretical example and the promising results on the use-case example indicate the feasibility of the proposed method. To further substantiate its applicability, additional use-case scenarios and a more detailed comparison to alternative approaches would be beneficial, e.g, a direct comparison to rejection sampling based approaches by \citet{toni2010} and \cite{didelot2011}. More recent approaches to compare to are given by \cite{vakilzadeh2018} who propose a method to estimate the model evidence via hierarchical state-space models, and by \cite{kacprzak2018qABC} who improve standard rejection sampling efficiency using the quantiles of the distance measure between simulated and observed data. Thus, a future project could be a systematic review of the performance of above mentioned methods and the method proposed here, on a tractable example problem and on a number of representative examples from different disciplines. 



\addtocontents{toc}{\vspace{2em}} 

\appendix 



\addtocontents{toc}{\vspace{2em}} 

\backmatter


\label{Bibliography}

\lhead{\emph{Bibliography}} 

\bibliographystyle{abbrvnat} 

\bibliography{main} 

\end{document}